\documentclass{article}

\PassOptionsToPackage{sort&compress}{natbib}
\usepackage{iclr2027_conference,times}
\setcitestyle{numbers,square,citesep={,}}

\usepackage{float}

\usepackage[utf8]{inputenc} % allow utf-8 input
\usepackage[T1]{fontenc}    % use 8-bit T1 fonts
\usepackage{hyperref}       % hyperlinks
\usepackage{url}            % simple URL typesetting
\usepackage{booktabs}       % professional-quality tables
\usepackage{amsfonts}       % blackboard math symbols
\usepackage{nicefrac}       % compact symbols for 1/2, etc.
\usepackage{microtype}      % microtypography
\usepackage{xcolor}         % colors

\definecolor{linkbrick}{RGB}{165,42,42}  % sections, equations, figures
\definecolor{citeteal}{RGB}{0,128,110}   % citations
\hypersetup{
    colorlinks=true,
    linkcolor=linkbrick,
    urlcolor=linkbrick,
    citecolor=citeteal,
}

\usepackage{tikz-cd}
\usepackage{comment}
\usepackage{booktabs}
\usepackage{multirow}
\usepackage{amsthm,amsmath,amssymb}
\usepackage{enumitem}

\usepackage{siunitx}
\newcommand{\R}{\mathbb{R}}
\newcommand{\C}{\mathbb{C}}

\newcommand{\N}{\mathbb{N}}

\newcommand{\Y}{\mathbb{Y}}

\newcommand{\lmax}{l_{\rm max}}

\newcommand{\Cxx}{C\texttt{++}}

\usepackage[ruled]{algorithm2e}
\usepackage{algorithmic}
\usepackage[utf8]{inputenc}
\usepackage{textcomp}
\usepackage{silence}
\DeclareUnicodeCharacter{00B2}{\ensuremath{^2}}
\DeclareUnicodeCharacter{208A}{\ensuremath{_+}}
\DeclareUnicodeCharacter{208B}{\ensuremath{_-}}
\DeclareUnicodeCharacter{2297}{\ensuremath{\otimes}}
\DeclareUnicodeCharacter{27E9}{\ensuremath{\rangle}}
\DeclareUnicodeCharacter{2207}{\ensuremath{\sum}}
\DeclareUnicodeCharacter{2211}{\ensuremath{\sum}}
\DeclareUnicodeCharacter{2032}{\ensuremath{'}}

\title{\centering
E3J: An Efficient and Open-Source Backend\\
for Euclidean Equivariant Operations\\
on GPU and TPU}

\author{%
  Olivier Peltre \\
  InstaDeep \\
  \And
  Armand Picard \\ 
  InstaDeep
  \And 
  Adrien Pichard \\ 
  InstaDeep \\
  \And 
  Miguel Bragança \\ 
  InstaDeep \\
  \And
  Luca Giacomoni\thanks{Contributed while at InstaDeep.} \\
  Prima Mente \\
  \And 
  Valentin Heyraud \\
  InstaDeep \\ 
  \And
  Zachary Weller-Davies \\
  InstaDeep \\ 
  \And
  Christoph Brunken \\ 
  InstaDeep \\
  \And
  Jules Tilly \\ 
  InstaDeep \\
}
\iclrfinalcopy
\begin{document}

\maketitle
\lhead{Preprint}

\begin{abstract}
We present {\tt e3j}, a fast Euclid-equivariance backend for geometric deep learning applications with JAX bindings for GPU and TPU. Leveraging both optimized CUDA and Pallas kernels and algorithmic improvements, the library achieves state-of-the-art throughput and runtime on both forward and backward paths. On a machine learning interatomic potential (MLIP) use case, it outperforms established backends, measuring up to 34\% speed-up over {\tt cuEquivariance} on water box NPT simulation using MACE, while remaining fully open source. {\tt e3j} achieves over 80 \% efficiency over the H100 maximum memory bandwidth on tensor product operations, and in many cases more than doubles throughput of message passing convolutions forward compared to previously available backends. In addition, with the release of dedicated Pallas TPU kernel, {\tt e3j} opens the possibility of large scale equivariant deep learning workloads on TPU architectures, which has so far been difficult to achieve. Our benchmarks show that {\tt e3j} also achieves over 80\% of a TPUv6e memory bandwidth, up to one order of magnitude more than {\tt e3nn\_jax}. 
The library is available on \href{https://github.com/instadeepai/e3j}{GitHub}, \href{https://pypi.org/project/e3j/}{PyPI} and is released under an open source Apache 2.0 license.
\end{abstract}

\section{Introduction}

% {\color{red} 
% One key novel feature of the {\tt e3j} library is the development of dedicated Pallas kernels for Euclidean Equivariant operation. To the best of our knowledge, no such kernels were published before, opening up scalable use of TPU for molecular dynamics workloads. To illustrate the benefits of such kernels, we present in figure \ref{fig:tp} the forward path of the tensor product comparing {\tt e3j} with {\tt e3nn\_jax}, which is, together with {\tt e3x}, one of the few libraries that can reliably be deployed on TPUs. 

% }

Euclid-equivariant operations are a core building block for many geometric deep learning applications. By providing exact equivariance guarantees, they have been shown to allow aggregation of high order geometric features in low data regime and without need for data augmentation \citep{Thomas-18, Nequip}. Their use has been explored across many fields that rely on geometric data, and in particular physical sciences, such as molecular dynamics \citep{MACE, MACE-OFF, Nequip, wood2026umafamilyuniversalmodels}, computational fluid dynamics (CFD) \citep{Kaszuba-25, Shankar23}, or electronic wavefunction and densities \citep{unke2021se3equivariantpredictionmolecularwavefunctions}.

In this work, we present {\tt e3j}, an open-source Euclid-equivariance backend providing all the core building blocks for E(3)-equivariant neural networks. The library is designed to facilitate constructions of these networks through a modular interface, but crucially is used as a means to integrate dedicated accelerated kernels into model architectures. To that end, the library includes kernels written both in Pallas and in CUDA, with the aim of deploying workloads across multiple types of architectures, including TPUs and GPUs. Namely, we present three sets of kernels backed by novel algorithms: 
\begin{itemize}
    \item \textbf{CUDA kernels}: Fully deterministic equivariant operations for GPU, focused on broad 
    compatibility across JAX versions through ahead-of-time (AOT) compilation.
    \item \textbf{Pallas GPU kernels}: Focused on performance, setting state-of-the-art throughput at time of writing in most regimes tested through just-in-time (JIT) compilation by the Mosaic compiler.
    Sacrifices determinism on the backward pass, compatibility with some older JAX versions, and support for 
    some operations (e.g. low number of channels).
    \item \textbf{Pallas TPU kernels}: Focused on delivering optimized throughput for TPUs.
\end{itemize}

We include and benchmark kernels specifically for both the equivariant tensor product, and for the full message passing convolution. In order to illustrate the capabilities of the library in realistic workloads, we focus on applications in the field of machine learning interatomic potentials (MLIPs) which are likely among the most relevant use case examples for the technology as (1) they require a complete absence of systematic bias from frame orientation to produce stable, long time molecular simulations, and (2) unlike other fields of application such as CFD, they rely on nearly perfectly equivariant training data. 

The library achieves throughputs comparable to those of closed source alternatives, for example reaching 2.7 TB/s on H100 GPU (81\% of memory bandwidth) on the forward tensor product operation, and 1.33 TB/s on TPUv6.2e (also 81\% of memory bandwidth). In order to illustrate the practical benefits of {\tt e3j}, we also benchmarked end-to-end integration within the MACE \citep{MACE,MACE-OFF} and NequIP \citep{Nequip} models, notably providing about 5x force inference speedup over {\tt e3nn} on TPU. On GPU, {\tt e3j} provides 2x speedup in simulations 
over the best open-source backend {\tt OpenEquivariance}, 
and 34\% speedup over the proprietary {\tt cuEquivariance} backend, reaching up to 64 thousands atoms (with $\sim 40$ connectivity degree) on MACE before overflowing the H100 memory.
Details of these results are presented in Section \ref{sec:results}. For theoretical details about the operations included in the library, readers can refer to Appendix \ref{apx:operations-scope}. The details of our novel algorithmic and methodological developments are presented in Appendix \ref{apx:kernel_description}. 

\section{Related Work}
    \paragraph{Engineering  work:} There has been considerable effort in building parameterized equivariant transforms accessible to the machine learning (ML) and scientific computing ecosystem. Generally, this consists in extending standard  ML frameworks such as JAX \citep{jax} and PyTorch \citep{pytorch} to benefit from their automatic differentiation (AD) support, so that equivariant building blocks can be seamlessly integrated within larger workflows on accelerated hardware such as GPUs and TPUs. 
These extensions may be defined either within the AD framework itself (JAX / PyTorch), or via the lower-level definition of ad-hoc primitives in the CUDA language\footnote{
    While the CUDA language from NVIDIA\textregistered\ extends 
    \Cxx~to let users write general purpose programs on their GPUs, 
    there is no low-level public kernel language for TPUs.
} \citep{CUDAC++}. The main examples of relevant libraries in this space are:

\begin{itemize}
\item {\tt  e3nn} \citep{e3nn_lib,e3nn_paper}, one of the first and most feature-rich Euclid equivariance libraries, consisting of two sibling packages written in Torch and JAX, respectively.
It has been used to construct widely used MLIP architectures such as MACE \citep{MACE,MACE-OFF} and NequIP \citep{Nequip}, which can learn to predict energies and forces at quantum-levels of
accuracy, thus disrupting the speed-accuracy compromise of MD simulations \citep{MLIP-paper}.

\item {\tt e3x} \citep{e3x}, an open source Euclid equivariance package written in JAX \citep{jax} which is slightly less flexible than ${\tt e3nn}$ due to its stricter data model, but offers efficient bilinear projections with cubic $L$ scaling \citep{Maennel2024}.

\item {\tt OpenEquivariance} \citep{openeq}, an open source CUDA kernel generation package with Torch and JAX bindings. It delivers state-of-the-art performance on focused operations (tensor product, message-passing convolution) with dedicated double-backward kernels to optimize training and Hessian inference. It is meant as a drop-in replacement for specific {\tt e3nn} operations 
on which it provides speedups of an order of magnitude.

\item {\tt cuEquivariance} ({\tt cuEq}), a proprietary NVIDIA\textregistered\ package with a toolset already as complete as {\tt e3nn}, providing efficient CUDA
kernels for tensor products and polynomial evaluation along with JAX and Torch bindings. {\tt cuEquivariance} is the most efficient backend, with an open-source API that however depends on closed-source kernels. 
\end{itemize}

The above list includes the most feature-complete packages viable for the construction of larger equivariant neural networks such as MLIPs. Efficient, near-optimal (open-source or closed-source) solutions therefore exist to compute Clebsch-Gordan tensor products 
on NVIDIA GPUs.
What best distinguishes {\tt e3j} from recent work is that {\tt e3j} provides a standalone and platform-agnostic API 
in JAX for GPU and TPU execution. 

\paragraph{Theoretical work:} Several alternative approaches to improve equivariant operations in geometric deep learning have come from the fundamental side. Naively a Clebsch--Gordan tensor product scales as $O(L^6)$, or as $O(L^5)$ if one exploits sparsity. Mathematical work has shown that cubic scaling with $L$ can be obtained for equivariant tensor products, though it is worth noting that these always come with compromise in terms of applicability or expressivity \citep{xie2025the}.

Some methods demonstrate cubic scaling with $L$ on arbitrary inputs. {\em Gaunt tensor product formulas}~\citep{Luo-24} are morally similar to performing element-wise products instead of discrete convolution via reciprocal Fourier transforms on the input and outputs. While the initial formulation of the Gaunt tensor product (GTP) however fails to fully reproduce the Clebsch-Gordan tensor product, as it does not incorporate anti-symmetric elements, limitations were remediated by a series of papers which formulate and incorporate an anti-symmetric counterpart called \textit{Vector Signal Tensor Product} (VSTP) \citep{xie2025the, xie2026asymptoticallyfastclebschgordantensor}, and later \citep{heyraud2026integralformulasvectorspherical, bochkarev2026fastcontractedclebschgordantensor} providing a closed form formula for efficient complete simulation of Clebsch-Gordan tensor products. Let us also mention the matrix tensor product formulation of~\citep{Maennel2024} which also has $O(L^3)$ scaling and is implemented in {\tt e3x}, providing order of magnitude speedups at $L=10$ over the equivalent {\tt e3nn} implementation. All of these methods however do result in automatic collapse of output multiplicity, a more efficient implementation which however does lead to a loss in expressivity \citep{xie2025the}.

Other methods specialize in delivering efficiency gains in the special case of a tensor product between an arbitrary feature vector of irreducible representations and equivariant features obtained by harmonic embeddings of an input vector. This particular case remains dominant in the literature, constituting one of the core building blocks of the original Tensor Field Network \citep{Thomas-18}, later used in NequIP and MACE. A first example is the $SO(2)$ convolution, presented in the equivariant Spherical Channel Network (eSCN) \citep{passaro2023escn} and notably used in eSEN \citep{fu2025learningsmoothexpressiveinteratomic} and UMA \citep{wood2026umafamilyuniversalmodels}, which defines a frame a reference from each edge and rotates the corresponding tensor product operands using the Wigner-D matrices. This results in harmonic features collapsing to $m=0$ across all $L$, removing the need to sum over $m$ indices of the harmonic features when performing the tensor product. This, combined with the additional use of symmetries in the Clebsch-Gordan tensor product achieves a cubic scaling in $L$.
A second example was presented in the E2former model \citep{li2026eformer} (later combined with $SO(2)$ convolutions by the same team \citep{huang2026e2formerv2ontheflyequivariantattention}) where the projected displacement vectors are replaced with the difference of projected input positions into harmonic features. Using a Binomial expansion, the authors show that one can construct a messaging passing block solely relying on node-wise tensor products. While message passing scaling remains unchanged, the number of tensor products (expected to be among the most costly operations) now scales with the number of nodes rather than the number of edges.

It is also worth noting that while theoretical work often focuses on tensor product scaling in $L$, most MLIP applications are in practice interested in the scaling in $N$ but at fixed $L$ (usually 2 or 3)\footnote{
    Applications such as signal processing or meteorology are in contrast
    interested in high maximal degree $L$. 
}.
Improving the scaling with $L$ typically requires clever re-parameterizations of equivariant features, and incurs a practical overhead that may not prove beneficial over an efficient implementation of the full Clebsch-Gordan tensor product in the small $L$ regime. 

We have conducted benchmarks of our backends against methods based on Gaunt tensor products and its extensions, {\tt e3x} methods and on $SO(2)$ convolution. These are presented in Sec. \ref{sec:results} and in more detail in Appendices \ref{apx:gaunt-comparison}, \ref{apx:e3x-comparison} and \ref{apx:so2-comparison} respectively.

\section{Results \label{sec:results}}

The {\tt e3j} package consists of a Python API targeting the JAX backend, alongside CUDA %\citep{CUDAC++} 
and Pallas kernel implementations for performance critical operations (tensor product, message-passing). 
The JAX framework \citep{jax} enables seamless integration within larger programs that can be \emph{just-in-time} (JIT) compiled to XLA (for Accelerated Linear Algebra), as are typically all recent MLIP model implementations.
In addition to atomic, module-wise benchmarks against reference E3 backends, we used the open-source {\tt mlip} library \citep{MLIP-paper} as reference and starting point to estimate the end-to-end speedups {\tt e3j} integration may provide in the MACE and NequIP equivariant architectures \citep{MACE,Nequip}. The details of the implementation method, algorithms, and differentiation rules can be found in Appendix \ref{apx:kernel_description}. 

It is worth noting that our JAX primitives bound to custom kernels are all made infinitely differentiable via recursive AD rules, and compatible with other higher-order JAX transforms ({\tt vmap, shard\_map,}...) to provide SPMD execution on GPU and TPU, an essential feature for MLIP training workflows on a large data scale. Architecture requirements for GPU and TPU however largely differ beyond that point, necessarily resulting in different algorithms. We have added a commentary on the GPU / TPU differences in appendix \ref{apx:hardware_concepts}.

In order to assess the performance of the library, we focus on two types of benchmarking (further discussion on the benchmark details can be found in Appendix \ref{apx:benchmark_details}): 

\begin{itemize}
    \item \textbf{Module specific:} We benchmark the performance critical components of any Euclidean equivariant library, namely: %namely the tensor product, and message passing convolution. 
    \begin{itemize}
        \item \textbf{Tensor product: } Bilinear coupling of latent equivariant features with static {\em Clebsch-Gordan coefficients}. In general these are performed per edge and per channel.
        \item \textbf{Message passing:} Aggregates the edge-wise features, usually computed through combination of harmonics projection, tensor product and linear or scalar mixing. The message passing aggregation tends to be the operational bottleneck once efficient tensor product operations are implemented.
    \end{itemize}

    \item \textbf{End-to-end:} To determine the overall relevance of the library in a complete workflow compared to alternatives, we also test two popular MLIP architectures, MACE \citep{MACE} and NequIP \citep{Nequip}. To maintain comparability, we connected the full benchmark in a fork of the open-source {\tt mlip} library \citep{MLIP-paper}, making the choice of backend the only variable differing in each run. Note that these benchmarks are for illustration only and that applicability of {\tt e3j} is not restricted to these two architectures (nor to MLIPs in general), a faithful and exhaustive comparison across all possible models and fields of application is not in scope of this work, for obvious reasons.
\end{itemize}

Experiments were conducted running JAX (v0.11.1) on an NVIDIA (R) H100-HBM3 GPU with 3.35 TB/s HBM, and on a TPUv6e Trillium with 1.64 TB/s HBM.
We benchmark seven different backends, five for GPU, and two for TPU. 
On GPU, we compare the following backends with our \textbf{CUDA} and \textbf{Pallas GPU} kernels: 
\begin{itemize}
    \item \textbf{cuEquivariance} (v0.11.1): Proprietary tensor product and message-passing convolution kernels of NVIDIA.
    \item \textbf{OpenEquivariance} (v0.7.0): We use JAX bindings to their equivalent tensor product and message-passing convolution kernels, which are JIT compiled from open-source CUDA kernels with NVRTC \citep{openeq}. Note that {\tt OpenEquivariance} provides two distinct convolution binaries, a deterministic one for graphs with edges sorted by receiver node index, and a non-deterministic one.
    \item \textbf{e3nn-jax} (v0.21.1): The reference {\tt e3nn-jax} library sets the performance threshold obtained by a pure JAX implementation,
    JIT compiled by XLA, but without any specific low level optimization. 
    It should therefore only be considered as an illustrative reference. 
    The {\tt e3nn} label uses the default half-precision for {\tt matmul}, 
    while {\tt e3nn\_f32} enforces single-precision in unit benchmarks.
\end{itemize}

On TPU, we compare our \textbf{Pallas TPU} kernels with XLA compiled \textbf{e3nn-jax} equivalent implementations of the full tensor product and message-passing convolution.
%Note that there are other available backends that we have excluded, most notably {\tt e3x} as direct comparison is not possible due to its stricter constant-multiplicity data model, without significant engineering and padding efforts to emulate the universal CGTP.

Additional benchmark results can be found in appendix \ref{apx:additional-benchmarks}, while comparisons with JAX implementations of cubically scaling 
algorithms (SO2, Gaunt/VSTP, {\tt e3x}) are grouped in appendix \ref{apx:cubically-scaling}. 
%Further details about the theoretical, algorithmic and engineering approaches used for each module can be found in %appendix \ref{sec:kernel_description}. 

\subsection{Unit benchmarks}

This section provides efficiency comparisons of the currently implemented {\tt e3j} kernels with available baselines in typical regimes. The tensor product operation is a full Clebsch-Gordan tensor product, satisfying the so-called 
{\em universal property}\footnote{
    Any bilinear map $b : (X, Y) \to Z$ factors through the tensor product space $X \otimes Y$ as a linear map 
    $\tilde{b}: X \otimes Y \to Z.$
} of tensor products. The message-passing convolution operation accumulates edge-wise tensor products on receiver nodes, and 
consists of a typical bottleneck in MLIP architectures \citep{EquiformerV2,passaro2023escn,fu2025learningsmoothexpressiveinteratomic}.

Runtimes were measured on XLA compiled, numerically equivalent implementations, averaging the fastest 20\% of 100 runs, and disabling the Python garbage collector using the {\tt timeit} module.  
The backward pass consists of the isolated, XLA compiled vector-jacobian product (VJP). 

Our metric of interest is throughput, i.e. the total amount of input/output bytes processed in the operation per unit of time. The peak throughput can be directly compared with the GPU / TPU maximum bandwidth to provide a meaningful speed relative to the device, independent of I/O size. 
In the backward pass benchmarks, only input primals, output cotangents and input cotangents 
were considered in the VJP accounting, excluding any saved residuals.

\subsubsection{Tensor Product}

\begin{figure}[h]
    \centering
    \includegraphics[width=0.95\linewidth]{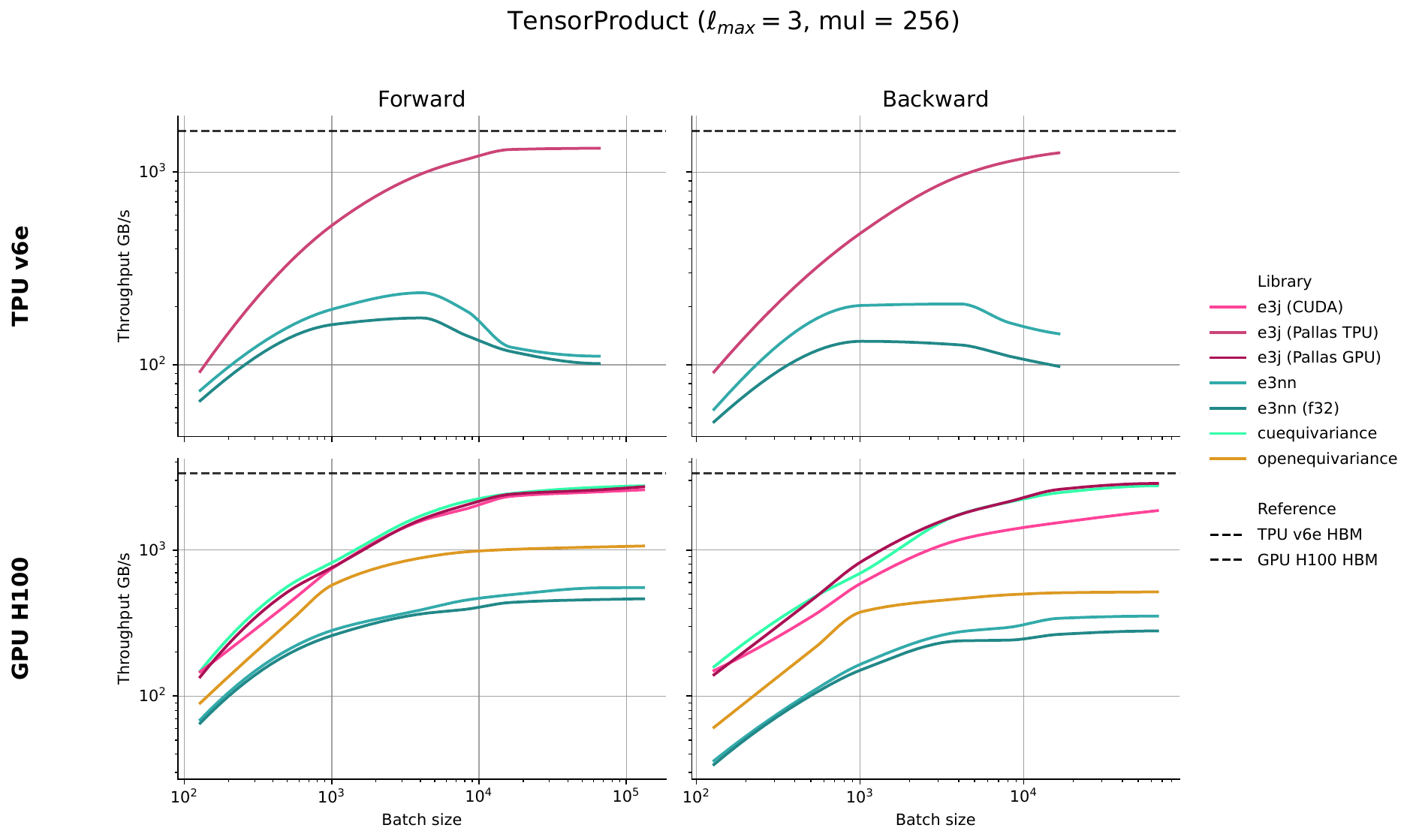}
    \caption{\textbf{Throughputs (GB/s) for the tensor product operation on TPU (top) and GPU (bottom)}. The {\tt e3j} CUDA implementation is compared against {\tt cuEq} (jax), and {\tt e3nn\_jax} (single precision and default half precision). The {\tt e3j} Pallas implementation is compared against the only available baseline {\tt e3nn\_jax} for 
    the universal ("full") Clebsch-Gordan tensor product.
    Results for $\ell_{max} = 2$ and $\ell_{max} = 3$ and multiplicities up to 256 channels, alongside the (a) theoretical maximum throughput of the NVIDIA H100 GPU ($\sim$ 3.35 TB/s) and (b) theoretical maximum throughput of a TPU v6e ($\sim$ 1.64 TB/s) on which the benchmarks were performed. When lines are not complete, it indicates that the library reached the memory bound.}
    \label{fig:tp}
\end{figure}

We developed dedicated tensor product kernels using CUDA, Pallas GPU and Pallas TPU, with results presented in Fig. \ref{fig:tp}. Our results show that the Clebsch-Gordan tensor product (CGTP) operation can reach up to 80\% HBM in the forward and backward passes on both GPU and TPU.  This means that CGTP in the small $L\leq 3$ regime is not a compute bottleneck in itself, while slightly larger degrees $L=4, 5\dots$ remain amenable to further engineering optimizations that were not prioritized at this time.
On GPU, our Pallas GPU kernel performs almost exactly on par with {\tt cuEquivariance}, while our CUDA kernel performs similarly on the forward pass and slightly below Pallas  GPU / {\tt cuEquivariance} on the backward pass. All three outpace {\tt OpenEquivariance}. On TPU our Pallas kernel achieves one order of magnitude higher throughput than the previously available {\tt e3nn\_jax}.

Note that our tensor product primitive is exposed as a generic bilinear coupling 
of operands with an arbitrary sparse COO array of coefficients. 
As presented in appendix \ref{apx:gaunt-comparison}, 
this allows us to further benchmark our kernels against Gaunt \citep{, Luo-24} and Vector Signal Tensor products \citep{xie2025the, xie2026asymptoticallyfastclebschgordantensor, heyraud2026integralformulasvectorspherical, bochkarev2026fastcontractedclebschgordantensor} on symmetric and skew-symmetric paths respectively.
We find (see figure \ref{fig:gaunt-comparison}) that for $L\leq 3$, our kernels are $2$ to $4$ times faster than the Gaunt tensor product on symmetric paths, with crossing at $L=6$ for the forward and $L=5$ for the backward. For the skew-symmetric paths, our kernels outperform the VSTP by a factor of $4$ to $6$ for $L\leq3$, and the crossing occurs one degree later due to the extra cost of the vector-valued operations in the VSTP. While we believe our implementations of GTP and VSTP to be efficient, dedicated kernel optimization of these operations could improve the relative results. 

\subsubsection{Message Passing Convolution}

\begin{figure}[hb]
    \centering
    \includegraphics[width=0.95\linewidth]{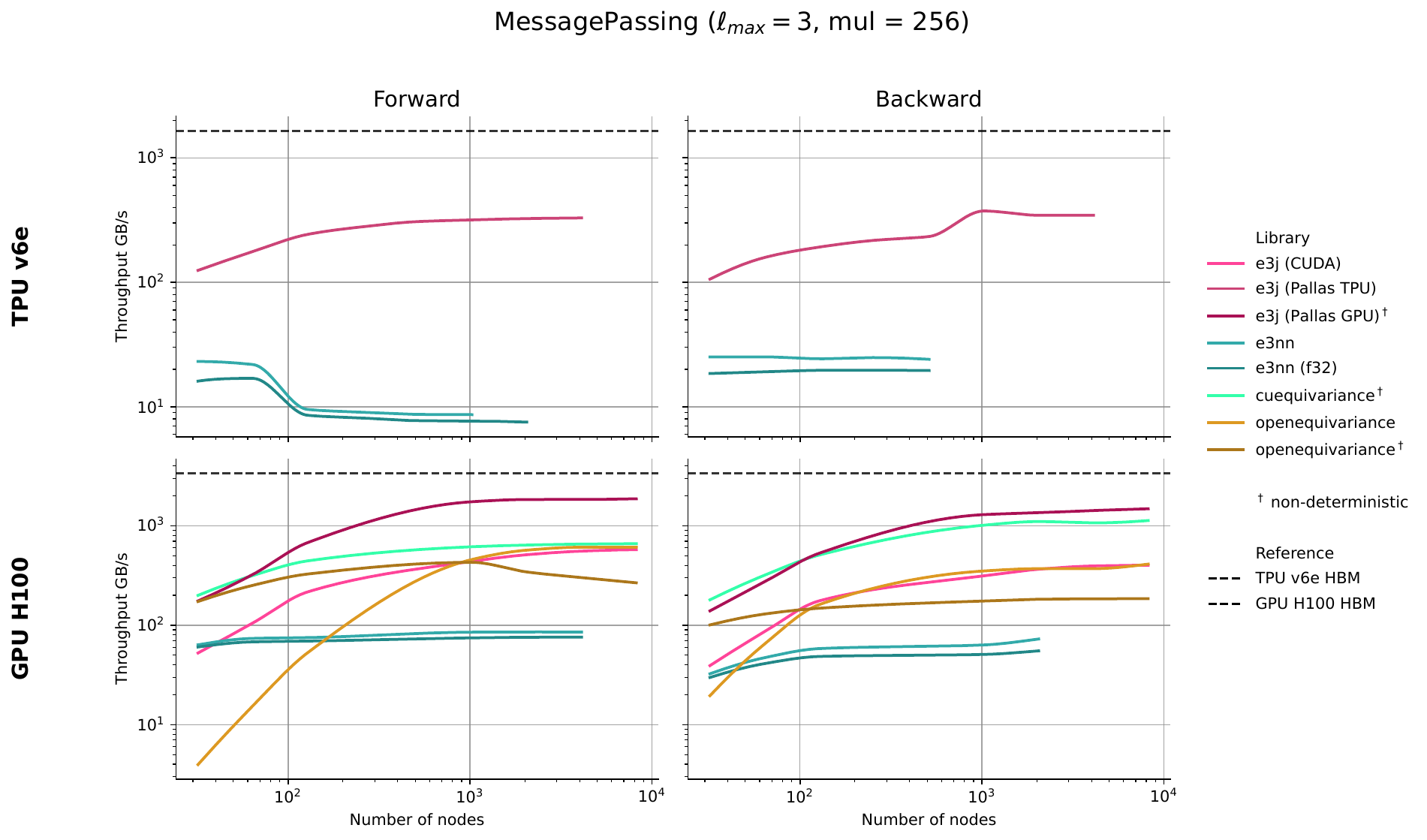}
    \caption{\textbf{Throughputs (GB/s) for the convolution operation on TPU (top) and GPU (bottom)}. 
    Here we present results for $\ell_{max} = 3$ 256 channels, alongside the theoretical maximum throughput of a TPU v6e Trillium (1.64 TB/s) and the theoretical maximum throughput of the NVIDIA H100 HBM3 GPU ($\sim$ 3.35 TB/s) on which the benchmarks were performed. When lines are not complete, it indicates that the library reached the memory bound of the device.}
    \label{fig:mp_conv}
\end{figure}

We present results for our three sets of Message Passing Convolution kernels: CUDA, Pallas GPU and Pallas TPU. On GPU, the Pallas GPU dominates the benchmarks, with over two times the throughput of {\tt cuEquivariance} in the forward pass under most settings, and a slightly higher throughput in the backward pass. It is worth noting that our Pallas GPU kernel is not (yet) compatible with channel counts lower than 128. The CUDA kernels are broadly on par with {\tt OpenEquivariance}, both having significantly lower throughput than Pallas GPU and {\tt cuEquivariance}. These results are presented in Tab. \ref{tab:conv-mul-scaling}. On TPU, our convolution kernels provide well over one order of magnitude speedups on the forward, and nearly one order of magnitude on the backward over {\tt e3nn-jax}. An overview of the results is presented in Fig. \ref{fig:mp_conv}. 

Our convolution kernels further provide the ability to skip padding edges, which typically all point to the same padding 
node\footnote{
    Given the graph topology is dynamic, while XLA compiled programs expect static shapes, 
    simulations typically add a single padding node to the atom list while managing a 
    fixed-size buffer of edges, only expanded when the real connecting edges overflow 
    the neighbor list (a rare event triggering re-compilation).
} in most JAX-based graph neural network frameworks \citep{jraph2020github}. 
This feature avoids the potentially significant overhead of aggregating messages on a padding node of unusually high valency, as illustrated by the simulation runtimes summarized in table \ref{table:npt_runtimes}. Implementation details are found in Appendix \ref{apx:kernel_description}.

\begin{table}
\centering \caption{\label{tab:conv-mul-scaling}
\textbf{Maximal convolution throughput (GB/s) by channel count on GPU.} 
End-to-end throughput is reported at the maximal power-of-two node count fitting on the 
NVIDIA H100 HBM3, with 45 average neighbors. This keeps the product of node count with 
channel count fixed to $2^{22}$ at 
$\ell_{max} = 2$, and to $2^{21}$ at $\ell_{max} = 3$ for fused kernels. 
The ${\tt e3nn}$ baseline is kept to illustrate the yield of a pure JAX implementation
of the same operation, but message materialization overflows memory at 
$2^{18}$ and $2^{17}$ respectively. The Pallas GPU kernel does not support channel 
counts below 128 yet (minimal block size imposed by Pallas).
}

\begin{tabular}{clcrrrrr}
\toprule
%\multicolumn{8}{c}{\bfseries Throughput by channel count (GB/s)} \\
%\midrule
& & & \multicolumn{5}{c}{Channels} \\
\cmidrule(l){4-8}
$\ell_{\max}$ & Implementation & Det. & \multicolumn{1}{c}{64} & \multicolumn{1}{c}{128} & \multicolumn{1}{c}{256} & \multicolumn{1}{c}{512} & \multicolumn{1}{c}{1024} \\
\specialrule{2\heavyrulewidth}{1.1ex}{0.9ex}
\multicolumn{8}{c}{\bfseries Forward pass} \\
\midrule
\multirow{6}{*}{2} & \texttt{e3nn} (f32)$^\dagger$ &  & \phantom{00}89\,{\scriptsize $\pm$\phantom{0}1} & \phantom{00}90\,{\scriptsize $\pm$\phantom{0}1} & \phantom{00}90\,{\scriptsize $\pm$\phantom{0}0} & \phantom{00}92\,{\scriptsize $\pm$\phantom{0}0} & \phantom{00}90\,{\scriptsize $\pm$\phantom{0}0} \\
 & \texttt{CuEquivariance} &  & \phantom{0}779\,{\scriptsize $\pm$\phantom{0}1} & \phantom{0}833\,{\scriptsize $\pm$15} & \phantom{0}786\,{\scriptsize $\pm$\phantom{0}4} & \phantom{0}807\,{\scriptsize $\pm$11} & \phantom{0}755\,{\scriptsize $\pm$\phantom{0}0} \\
 & \texttt{OpenEquivariance} & \checkmark & \bfseries \phantom{0}871\,{\scriptsize $\pm$\phantom{0}3} & 1092\,{\scriptsize $\pm$\phantom{0}8} & \phantom{0}760\,{\scriptsize $\pm$\phantom{0}8} & \phantom{0}631\,{\scriptsize $\pm$\phantom{0}5} & \phantom{0}652\,{\scriptsize $\pm$\phantom{0}7} \\
 & \texttt{OpenEquivariance} &  & \phantom{0}348\,{\scriptsize $\pm$\phantom{0}1} & \phantom{0}343\,{\scriptsize $\pm$\phantom{0}1} & \phantom{0}370\,{\scriptsize $\pm$\phantom{0}1} & \phantom{0}354\,{\scriptsize $\pm$\phantom{0}2} & \phantom{0}443\,{\scriptsize $\pm$\phantom{0}1} \\
 & \texttt{e3j} (CUDA) & \checkmark & \phantom{0}625\,{\scriptsize $\pm$\phantom{0}2} & \phantom{0}955\,{\scriptsize $\pm$\phantom{0}4} & \phantom{0}913\,{\scriptsize $\pm$\phantom{0}4} & \phantom{0}956\,{\scriptsize $\pm$\phantom{0}2} & \phantom{0}880\,{\scriptsize $\pm$\phantom{0}5} \\
 & \texttt{e3j} (Pallas GPU) & \checkmark & \phantom{00}--\,{\scriptsize \phantom{$\pm$00}} & \bfseries 1621\,{\scriptsize $\pm$\phantom{0}9} & \bfseries 1669\,{\scriptsize $\pm$12} & \bfseries 1716\,{\scriptsize $\pm$29} & \bfseries 1871\,{\scriptsize $\pm$41} \\
\midrule
\multirow{6}{*}{3} & \texttt{e3nn} (f32)$^\dagger$ &  & \phantom{00}73\,{\scriptsize $\pm$\phantom{0}0} & \phantom{00}72\,{\scriptsize $\pm$\phantom{0}1} & \phantom{00}75\,{\scriptsize $\pm$\phantom{0}0} & \phantom{00}75\,{\scriptsize $\pm$\phantom{0}1} & \phantom{00}73\,{\scriptsize $\pm$\phantom{0}1} \\
 & \texttt{CuEquivariance} &  & \phantom{0}829\,{\scriptsize $\pm$11} & 1072\,{\scriptsize $\pm$\phantom{0}3} & \phantom{0}662\,{\scriptsize $\pm$\phantom{0}1} & \phantom{0}632\,{\scriptsize $\pm$\phantom{0}1} & \phantom{0}629\,{\scriptsize $\pm$\phantom{0}0} \\
 & \texttt{OpenEquivariance} & \checkmark & \bfseries \phantom{0}856\,{\scriptsize $\pm$\phantom{0}2} & \phantom{0}709\,{\scriptsize $\pm$\phantom{0}1} & \phantom{0}612\,{\scriptsize $\pm$\phantom{0}1} & \phantom{0}565\,{\scriptsize $\pm$\phantom{0}2} & \phantom{0}347\,{\scriptsize $\pm$\phantom{0}5} \\
 & \texttt{OpenEquivariance} &  & \phantom{0}265\,{\scriptsize $\pm$\phantom{0}1} & \phantom{0}268\,{\scriptsize $\pm$\phantom{0}1} & \phantom{0}303\,{\scriptsize $\pm$\phantom{0}1} & \phantom{0}323\,{\scriptsize $\pm$\phantom{0}0} & \phantom{0}303\,{\scriptsize $\pm$\phantom{0}4} \\
 & \texttt{e3j} (CUDA) & \checkmark & \phantom{0}389\,{\scriptsize $\pm$\phantom{0}0} & \phantom{0}569\,{\scriptsize $\pm$\phantom{0}2} & \phantom{0}573\,{\scriptsize $\pm$\phantom{0}1} & \phantom{0}562\,{\scriptsize $\pm$\phantom{0}2} & \phantom{0}538\,{\scriptsize $\pm$\phantom{0}2} \\
 & \texttt{e3j} (Pallas GPU) & \checkmark & \phantom{00}--\,{\scriptsize \phantom{$\pm$00}} & \bfseries 1818\,{\scriptsize $\pm$\phantom{0}6} & \bfseries 1859\,{\scriptsize $\pm$14} & \bfseries 2074\,{\scriptsize $\pm$14} & \bfseries 2067\,{\scriptsize $\pm$19} \\
\specialrule{2\heavyrulewidth}{1.1ex}{0.9ex}
\multicolumn{8}{c}{\bfseries Backward pass} \\
\midrule
\multirow{6}{*}{2} & \texttt{e3nn} (f32)$^\dagger$ &  & \phantom{00}63\,{\scriptsize $\pm$\phantom{0}1} & \phantom{00}65\,{\scriptsize $\pm$\phantom{0}1} & \phantom{00}64\,{\scriptsize $\pm$\phantom{0}0} & \phantom{00}62\,{\scriptsize $\pm$\phantom{0}2} & \phantom{00}60\,{\scriptsize $\pm$\phantom{0}1} \\
 & \texttt{CuEquivariance} &  & \bfseries 1245\,{\scriptsize $\pm$\phantom{0}3} & \bfseries 1287\,{\scriptsize $\pm$\phantom{0}2} & 1263\,{\scriptsize $\pm$18} & 1264\,{\scriptsize $\pm$22} & 1248\,{\scriptsize $\pm$15} \\
 & \texttt{OpenEquivariance} & \checkmark & \phantom{0}767\,{\scriptsize $\pm$\phantom{0}7} & \phantom{0}627\,{\scriptsize $\pm$\phantom{0}4} & \phantom{0}574\,{\scriptsize $\pm$\phantom{0}3} & \phantom{0}573\,{\scriptsize $\pm$\phantom{0}4} & \phantom{0}564\,{\scriptsize $\pm$\phantom{0}4} \\
 & \texttt{OpenEquivariance} &  & \phantom{0}806\,{\scriptsize $\pm$\phantom{0}7} & \phantom{0}598\,{\scriptsize $\pm$\phantom{0}3} & \phantom{0}504\,{\scriptsize $\pm$\phantom{0}2} & \phantom{0}233\,{\scriptsize $\pm$\phantom{0}1} & \phantom{0}205\,{\scriptsize $\pm$\phantom{0}1} \\
 & \texttt{e3j} (CUDA) & \checkmark & \phantom{0}446\,{\scriptsize $\pm$\phantom{0}1} & \phantom{0}709\,{\scriptsize $\pm$\phantom{0}2} & \phantom{0}638\,{\scriptsize $\pm$\phantom{0}5} & \phantom{0}560\,{\scriptsize $\pm$\phantom{0}3} & \phantom{0}623\,{\scriptsize $\pm$\phantom{0}4} \\
 & \texttt{e3j} (Pallas GPU) &  & \phantom{00}--\,{\scriptsize \phantom{$\pm$00}} & 1267\,{\scriptsize $\pm$15} & \bfseries 1292\,{\scriptsize $\pm$\phantom{0}9} & \bfseries 1319\,{\scriptsize $\pm$10} & \bfseries 1328\,{\scriptsize $\pm$30} \\
\midrule
\multirow{6}{*}{3} & \texttt{e3nn} (f32)$^\dagger$ &  & \phantom{00}56\,{\scriptsize $\pm$\phantom{0}0} & \phantom{00}53\,{\scriptsize $\pm$\phantom{0}1} & \phantom{00}52\,{\scriptsize $\pm$\phantom{0}1} & \phantom{00}51\,{\scriptsize $\pm$\phantom{0}1} & \phantom{00}27\,{\scriptsize $\pm$\phantom{0}1} \\
 & \texttt{CuEquivariance} &  & \bfseries 1059\,{\scriptsize $\pm$\phantom{0}8} & 1104\,{\scriptsize $\pm$\phantom{0}3} & 1119\,{\scriptsize $\pm$12} & 1118\,{\scriptsize $\pm$22} & 1128\,{\scriptsize $\pm$29} \\
 & \texttt{OpenEquivariance} & \checkmark & \phantom{0}451\,{\scriptsize $\pm$\phantom{0}2} & \phantom{0}458\,{\scriptsize $\pm$\phantom{0}1} & \phantom{0}411\,{\scriptsize $\pm$\phantom{0}9} & \phantom{0}324\,{\scriptsize $\pm$\phantom{0}6} & \phantom{0}253\,{\scriptsize $\pm$\phantom{0}2} \\
 & \texttt{OpenEquivariance} &  & \phantom{0}470\,{\scriptsize $\pm$\phantom{0}2} & \phantom{0}190\,{\scriptsize $\pm$\phantom{0}0} & \phantom{0}185\,{\scriptsize $\pm$\phantom{0}1} & \phantom{0}151\,{\scriptsize $\pm$\phantom{0}0} & \phantom{0}154\,{\scriptsize $\pm$\phantom{0}1} \\
 & \texttt{e3j} (CUDA) & \checkmark & \phantom{0}310\,{\scriptsize $\pm$\phantom{0}0} & \phantom{0}423\,{\scriptsize $\pm$\phantom{0}2} & \phantom{0}402\,{\scriptsize $\pm$\phantom{0}2} & \phantom{0}317\,{\scriptsize $\pm$\phantom{0}1} & \phantom{0}183\,{\scriptsize $\pm$\phantom{0}0} \\
 & \texttt{e3j} (Pallas GPU) &  & \phantom{00}--\,{\scriptsize \phantom{$\pm$00}} & \bfseries 1407\,{\scriptsize $\pm$\phantom{0}3} & \bfseries 1432\,{\scriptsize $\pm$18} & \bfseries 1483\,{\scriptsize $\pm$22} & \bfseries 1599\,{\scriptsize $\pm$12} \\
\bottomrule
\end{tabular}
\end{table}

We also compare the performance of our CUDA and Pallas GPU kernels against the $SO(2)$ convolution \citep{passaro2023escn}. Our convolution kernels are $4$ to $6$ times faster than a JAX-based $SO(2)$ convolution up to $L=3$ (see \citep{MLIP-paper} for implementation details), while curves cross at $L=4$ for the forward, and $L=5$ for the backward (see figure \ref{fig:so2-comparison}) when force-collapsing the multiplicity to maintain an equivalent level of expressivity. Further details and plots can be found in Appendix \ref{apx:so2-comparison}. It is worth noting that the authors of UMA \citep{wood2026umafamilyuniversalmodels}, also produced a Triton based kernel optimization for the convolution which reduces runtime by a third though fixed for $L=2$. Assuming this improvement was ported into the JAX version it would still remain $\sim 6$ times slower than our Pallas GPU convolution kernel. 

\subsection{End-to-end benchmarks}

In order to assess the library in a practical setting, we performed end-to-end MLIP profiling and benchmarks of MACE and NequIP architectures. We compare our implementation with the $\tt{e3nn\_jax}$ baseline of the {\tt mlip} library \citep{MLIP-paper}, chosen for its unified and fast integration in downstream workflows (batched inference and simulations or relaxations), and to make sure that the backends are directly comparable.
For each architecture, we connected {\tt e3j}, {\tt cuEquivariance} and {\tt OpenEquivariance} as numerically equivalent message-passing backends.
% making sure to avoid any runtime adapter overhead.

Although MLIP models have linear $O(N)$ complexity in the total number of atoms $N$, the message-passing step scales with the number of edges and the average connectivity of the graph is typically larger than 40. For our MACE model (a) variant, 
message materialization may effectively bound achievable system sizes to around 13 thousand atoms before reaching memory overflows, 
while fused message-passing makes it possible to process up to 
64 thousand atoms on a single NVIDIA{\textregistered} H100 GPU with 80 GB of global memory.

\begin{figure}[ht]
    \centering
    \includegraphics[width=0.9\linewidth]{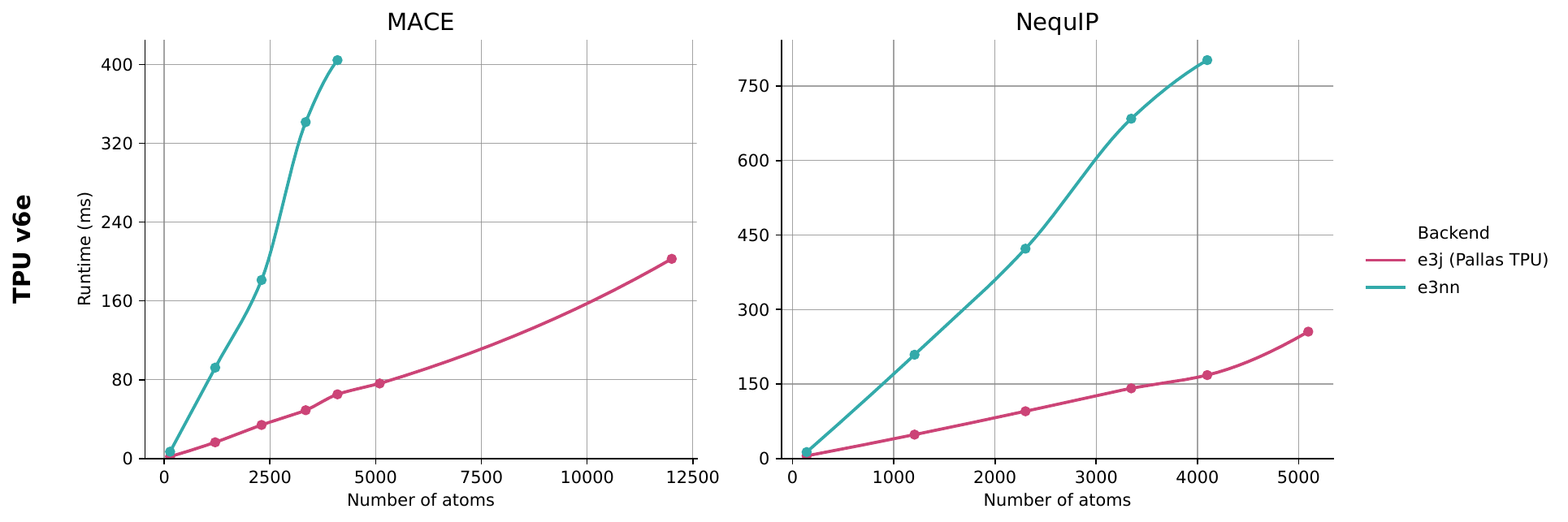}

    \caption{ \label{table:end_to_end_TPU}
    \textbf{Runtime (ms) for end-to-end force inference of MACE and NequIP on TPU}.
    Inference is performed through integration of additional convolution backends for the {\tt mlip} library \citep{MLIP-paper} and run on real protein systems with {5\AA} cutoff. 
    Hyperparameters can be found in table \ref{table:hparams}: 
    MACE (a) (correlation 2) has 2 layers and NequIP has 5 layers, both have 128 channels.}
    \label{fig:e2e_tpu}
\end{figure}

\begin{figure}[ht]
    \centering
    \includegraphics[width=0.9\linewidth]{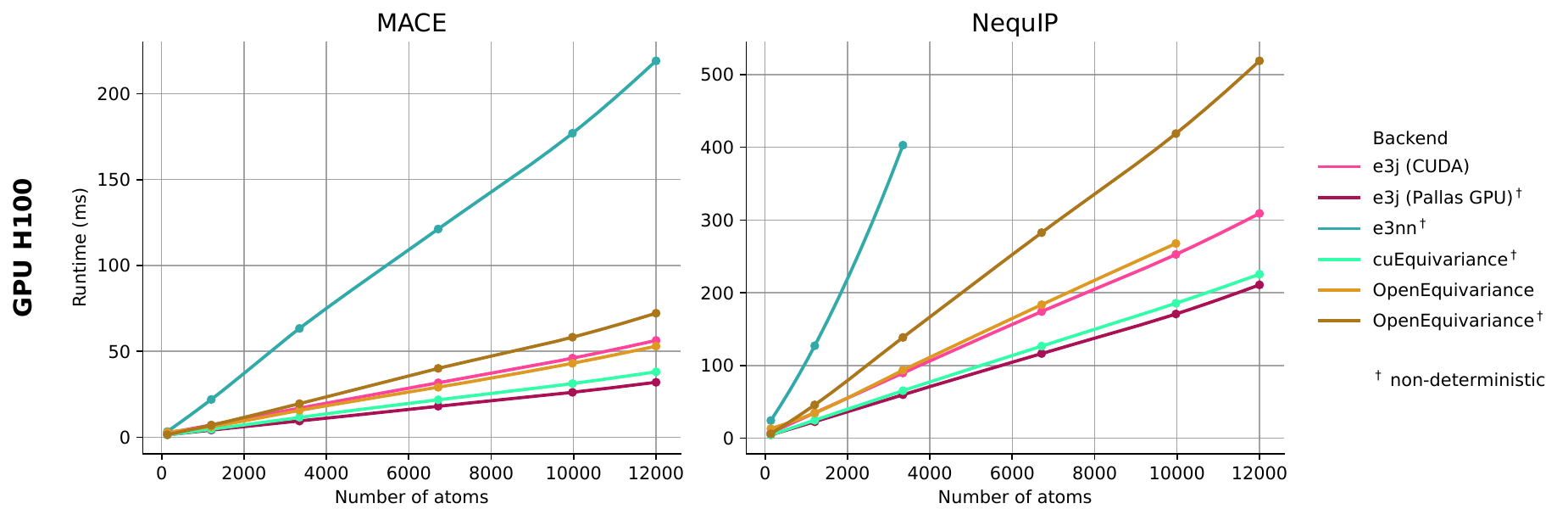}

    \caption{\label{table:end_to_end_GPU}
    \textbf{Runtime (ms) for end-to-end force inference of MACE and NequIP on GPU}.
    Inference is performed through integration of additional convolution backends for the {\tt mlip} library \citep{MLIP-paper} and run on real protein systems with {5\AA} cutoff. 
    Hyperparameters can be found in table \ref{table:hparams}: MACE (a) (correlation 2) has 2 layers and NequIP has 5 layers, 
    both have 128 channels.
    }
    \label{fig:e2e_gpu}
\end{figure}

Our results for MACE and NequIP are detailed in figures \ref{fig:e2e_tpu} and \ref{fig:e2e_gpu}. 
It is worth noting that while not tested in this paper, the library can also be deployed across a number of alternative MLIP architectures such as GRACE \citep{Lysogorskiy2026} and Equiformer / E2Former \citep{Equiformer, EquiformerV2, liao2026equiformerv3scalingefficientexpressive, li2026eformer}. 
%Traces obtained with the JAX profiler give precious insights into which equivariant operations are actual %bottlenecks and deserve to be optimized. 

\begin{table}[h]
  \centering
  \caption{\textbf{End-to-end NPT performance of a MACE model on a 25\AA\, water box (GPU).}
  Runtimes are reported for 100ps long NPT simulations with a Monte Carlo barostat, 
  Hyperparameters are from the MACE (a) variant of table \ref{table:hparams},
  notably ${\tt correlation}=2$ and ${\tt node\_symmetry}=2$.
  Only the convolution block is dispatched to dedicated kernels matching 
  the reference implementation numerically.
  %without any runtime adapter overhead. 
  The initial structure (solvated 2-methyl-butane, equilibrated with a classical force field) consists of 1503 atoms and 10\% initial edge padding (83,325 static edge count).
}
    \vspace{0.3em}
  \label{table:npt_runtimes}
  
% 25 A water box, npt_mc_langevin, 100 ps in 100 episodes of 1 ps.
% 1503 atoms; 78,141 real edges mean, 80,868 peak, against a fixed
% capacity of 83,325 (93.8% non-padding).
% No episode overflowed: 2.95% headroom at the peak, so the tightest
% capacity that would have survived is the peak itself, 80,868.
% jax 0.11.1; cuEquivariance 0.11.1, OpenEquivariance 0.7.0,
% e3j pinned at branch baselines-mgpu-b11.
% +- is the population sigma over the 99 production episodes.
\begin{tabular}{lccrr}
  \toprule
  Backend & Deterministic & Open-source & ms/step & ns/day \\
  \midrule
  \texttt{e3j} (CUDA) & \checkmark & \checkmark & \textbf{8.821}\,{\scriptsize $\pm$ 0.045} & \textbf{9.80}\,{\scriptsize $\pm$ 0.05} \\
  \texttt{OpenEquivariance} & \checkmark & \checkmark & 16.137\,{\scriptsize $\pm$ 1.063} & 5.38\,{\scriptsize $\pm$ 0.35} \\
  \midrule
  \texttt{cuEquivariance} &  &  & 6.586\,{\scriptsize $\pm$ 0.013} & 13.12\,{\scriptsize $\pm$ 0.03}\\
  \texttt{e3j} (Pallas GPU) &  & \checkmark & \textbf{4.906}\,{\scriptsize $\pm$ 0.014} & \textbf{17.61}\,{\scriptsize $\pm$ 0.05} \\
  \texttt{OpenEquivariance} &  & \checkmark & 10.041\,{\scriptsize $\pm$ 0.040} & 8.60\,{\scriptsize $\pm$ 0.03} \\
  \texttt{e3nn} &  & \checkmark & 68.298\,{\scriptsize $\pm$ 0.063} & 1.27\,{\scriptsize $\pm$ 0.00} \\
  \bottomrule
\end{tabular}
\end{table}

\begin{comment}
In NequIP, most of the runtime is spent in the message-passing block, since it contains a tensor product of edge features whose batch axis is about 40x larger than the number of nodes. This is also the case of MACE with $\tt{correlation = 2}$, i.e. when only equivariant squares of node features are computed. However the MACE architecture has the peculiarity of computing many-body features by iterating equivariant powers of node features in its $\tt{SymmetricContraction}$ block, and the situation would be
very different with $\tt{correlation = 3}$ as the cubic power begins to dominate the runtime. 
See Appendix \ref{apx:symmetric-contraction} for experiments including the 
{\tt SymmetricContraction} kernel of {\tt cuEquivariance}. 
\end{comment}

One point to note is that in all MACE benchmarks, the symmetric contraction relies on the CUDA tensor product kernel of {\tt e3j} with channel-mixing mode {\tt MAP}. 
This helps enforce numerical consistency and allowed 
us to run stable simulations from a single trained checkpoint. 
It also has a relatively small impact with the {\tt correlation = 2} results presented here. See Appendix \ref{apx:symmetric-contraction} for experiments at {\tt correlation = 3} involving the proprietary {\tt cuEquivariance} kernel for the symmetric contraction step and more discussion.

Paradoxically, table \ref{table:end_to_end_GPU} shows {\tt OpenEquivariance} 
leads to slower simulations with the deterministic convolution kernel 
(see figure \ref{fig:mp_conv}). This gap increases dramatically with the number of padding 
edges (above 30 ms/step with 25\% padding), indicating their kernel hangs waiting for 
the slowest block accumulating messages on the single padding node. 
Our kernels flag padding edges so that work on these edges can be skipped, avoiding this overhead. 

%\section{Analysis}
%\input{40_analysis}

\section{Discussion}
Equivariant architectures have often been criticized for being computationally heavy, with learned equivariance often being put forward as an efficient inference time alternative. Multiple architectures have recently moved away from the strict inductive bias in favor of transformer architectures \citep{qu2026recipescalableattentionbasedmlips, elhag2026learninginteratomicpotentialsexplicit}, using the advantageous engineering of the transformer architecture as motivation for the change. In this paper we show that with proper engineering, full Clebsch-Gordan tensor product and associated message passing convolutions can be efficiently implemented on both GPUs and TPUs, reaching over 80\% of HBM throughput. This should provide a more even playing field when comparing highly engineered transformer architectures with equivariant networks. 

It is worth noting that while dedicated kernels do improve the efficiency of the Clebsch-Gordan coefficients, and appear to be the best performing operation for $L<5$, they remain at a disadvantage in terms of scaling in $L$ compared to other methods, in particular the $SO(2)$ convolution. Full treatment of the tensor product however has the benefit of being applicable to arbitrary feature vector pairs, and not solely in combination with harmonic features. A fair comparison would also require for $SO(2)$ convolutions to also receive dedicated engineering, an effort that has begun with the second version of UMA \citep{wood2026umafamilyuniversalmodels} but that remains poorly explored by the community. 

One key point to note however, is that the achieved throughput remains below maximum bandwidth of the H100 GPU and v6e TPU, suggesting further improvements may be possible. We believe the release of \texttt{e3j} as an open-source library will prove a valuable new starting point for the community to continue optimizing these operations on current and future hardware.

\subsection*{AI use statement}

In this work, we used generative AI tools to help implement methods.
We have not used generative AI tools to help develop theoretical models or conceptual frameworks, formulate mathematical claims, propose or refine hypotheses, design or provide feedback on research methodology or experiments, assist with translation, support qualitative and thematic data analysis, interpret results,
and the rest of the required disclosure tasks (generate synthetic data sets, help develop theoretical models or conceptual frameworks, provide critical ingredients for proving mathematical claims, assist in the writing of proofs, clean and reformat dataset)
are not applicable to this work.
Additionally, we used generative AI tools to modify scientific figures and tables, create or edit software code. 
We have reviewed all AI-assisted work. 
LLM-generated code was reviewed by more than 2 authors and 
tested for correctness. 
We take responsibility for the final content of this work,
including text, claims or artifacts produced with the aid of generative AI.

\section*{Acknowledgments}

This work was supported by Cloud TPUs from Google’s TPU Research Cloud (TRC). 

We would like to express our gratitude to
Sébastien B. and Oliver B. for encouragement and support
in the early development of the library, and to warmly thank Marco C. for his continued 
assistance in the MLIP integration effort.

%%%%%%%%%%%%%%%%%%%%%%%%%%%%%%%%%%%%%%%%%%%%%%%%%%%%%%%%%%%%

%\bibliographystyle{plainnat}
\bibliographystyle{unsrtnat}
\bibliography{main}

@book{Wigner-59,
    title={{Group Theory And Its Application to the Quantum Mechanics of Atomic Spectra}},
    subtitle={{And Its Application to the Quantum Mechanics of Atomic Spectra}},
    author={Eugene Wigner},
    year={1959},
    publisher={Elsevier Science},
}

@article{Kaszuba-25,
    title = {{Implicit modeling of equivariant tensor basis with Euclidean turbulence closure neural network}},
    author = {Kaszuba, Grzegorz and Krakowski, Tomasz and Ziegler, Bartosz and Jaszkiewicz, Andrzej and Sankowski, Piotr},
    year = {2025},
    month = {02},
    pages = {},
    volume = {37},
    journal = {Physics of Fluids},
    doi = {10.1063/5.0249490}
}

@misc{Hall-2000,
      title={{An Elementary Introduction to Groups and Representations}}, 
      author={Brian C. Hall},
      year={2000},
      eprint={math-ph/0005032},
      archivePrefix={arXiv},
      primaryClass={math-ph},
      url={https://arxiv.org/abs/math-ph/0005032}, 
}

@misc{jax,
  author = {James Bradbury and Roy Frostig and Peter Hawkins and Matthew James Johnson and Chris Leary and Dougal Maclaurin and George Necula and Adam Paszke and Jake Vander{P}las and Skye Wanderman-{M}ilne and Qiao Zhang},
  title = {{JAX}: composable transformations of {P}ython+{N}um{P}y programs},
  url = {http://github.com/jax-ml/jax},
  version = {0.3.13},
  year = {2018}
}

@inproceedings{pytorch,
  title={Automatic differentiation in PyTorch},
  author={Paszke, Adam and Gross, Sam and Chintala, Soumith and Chanan, Gregory and Yang, Edward and DeVito, Zachary and Lin, Zeming and Desmaison, Alban and Antiga, Luca and Lerer, Adam},
  year={2017},
  booktitle={NIPS-W}
}

@manual{CUDAC++,
    title = {CUDA C++ Programming Guide},
    author={NVIDIA Corporation},
    url={https://docs.nvidia.com/cuda/archive/12.8.1/cuda-c-programming-guide/index.html},
    version = {12.8.1},
    key = {NVIDIA},
    year={2025}
}

@misc{MLIP-paper,
      title={Machine Learning Interatomic Potentials: library for efficient training, model development and simulation of molecular systems}, 
      author={Christoph Brunken and Olivier Peltre and Heloise Chomet and Lucien Walewski and Manus McAuliffe and Valentin Heyraud and Solal Attias and Martin Maarand and Yessine Khanfir and Edan Toledo and Fabio Falcioni and Marie Bluntzer and Silvia Acosta-Gutiérrez and Jules Tilly},
      year={2025},
      eprint={2505.22397},
      archivePrefix={arXiv},
      primaryClass={physics.chem-ph},
      url={https://arxiv.org/abs/2505.22397}, 
}

@misc{openeq,
      title={An Efficient Sparse Kernel Generator for O(3)-Equivariant Deep Networks}, 
      author={Vivek Bharadwaj and Austin Glover and Aydin Buluc and James Demmel},
      year={2025},
      eprint={2501.13986},
      archivePrefix={arXiv},
      primaryClass={cs.LG},
      url={https://arxiv.org/abs/2501.13986}, 
}

@misc{unke2021se3equivariantpredictionmolecularwavefunctions,
      title={SE(3)-equivariant prediction of molecular wavefunctions and electronic densities}, 
      author={Oliver T. Unke and Mihail Bogojeski and Michael Gastegger and Mario Geiger and Tess Smidt and Klaus-Robert Müller},
      year={2021},
      eprint={2106.02347},
      archivePrefix={arXiv},
      primaryClass={physics.chem-ph},
      url={https://arxiv.org/abs/2106.02347}, 
}

@misc{MACE,
      title={{MACE: Higher Order Equivariant Message Passing Neural Networks for Fast and Accurate Force Fields}}, 
      author={Ilyes Batatia and Dávid Péter Kovács and Gregor N. C. Simm and Christoph Ortner and Gábor Csányi},
      year={2023},
      eprint={2206.07697},
      archivePrefix={arXiv},
      primaryClass={stat.ML},
      url={https://arxiv.org/abs/2206.07697}, 
}

@misc{MACE-OFF,
      title={{MACE-OFF: Transferable Short Range Machine Learning Force Fields for Organic Molecules}}, 
      author={Dávid Péter Kovács and J. Harry Moore and Nicholas J. Browning and Ilyes Batatia and Joshua T. Horton and Yixuan Pu and Venkat Kapil and William C. Witt and Ioan-Bogdan Magdău and Daniel J. Cole and Gábor Csányi},
      year={2025},
      eprint={2312.15211},
      archivePrefix={arXiv},
      primaryClass={physics.chem-ph},
      url={https://arxiv.org/abs/2312.15211}, 
}

@misc{Thomas-18,
      title={Tensor field networks: Rotation- and translation-equivariant neural networks for 3D point clouds}, 
      author={Nathaniel Thomas and Tess Smidt and Steven Kearnes and Lusann Yang and Li Li and Kai Kohlhoff and Patrick Riley},
      year={2018},
      eprint={1802.08219},
      archivePrefix={arXiv},
      primaryClass={cs.LG},
      url={https://arxiv.org/abs/1802.08219}, 
}

@article{Nequip,
    title = {{E(3)-equivariant graph neural networks for data-efficient and accurate interatomic potentials}},
    author = {Batzner,  Simon and Musaelian,  Albert and Sun,  Lixin and Geiger,  Mario and Mailoa,  Jonathan P. and Kornbluth,  Mordechai and Molinari,  Nicola and Smidt,  Tess E. and Kozinsky,  Boris},
  journal = {Nature Communications},
  volume = {13},
  ISSN = {2041-1723},
  url = {http://dx.doi.org/10.1038/s41467-022-29939-5},
  DOI = {10.1038/s41467-022-29939-5},
  number = {1},
  publisher = {Springer Science and Business Media LLC},
  
  year = {2022},
  month = may 
}

@inproceedings{Equiformer,
title={Equiformer: Equivariant Graph Attention Transformer for 3D Atomistic Graphs},
author={Yi-Lun Liao and Tess Smidt},
booktitle={The Eleventh International Conference on Learning Representations },
year={2023},
url={https://openreview.net/forum?id=KwmPfARgOTD}
}

@inproceedings{EquiformerV2,
title={EquiformerV2: Improved Equivariant Transformer for Scaling to Higher-Degree Representations},
author={Yi-Lun Liao and Brandon M Wood and Abhishek Das and Tess Smidt},
booktitle={The Twelfth International Conference on Learning Representations},
year={2024},
url={https://openreview.net/forum?id=mCOBKZmrzD}
}

@misc{e3nn_paper,
    doi = {10.48550/ARXIV.2207.09453},
    url = {https://arxiv.org/abs/2207.09453},
    author = {Geiger, Mario and Smidt, Tess},
    title = {e3nn: Euclidean Neural Networks},
    publisher = {arXiv},
    year = {2022},
    copyright = {Creative Commons Attribution 4.0 International}
}

@misc{e3nn_lib,
    author = {Mario Geiger and
              Tess Smidt and
              Alby M. and
              Benjamin Kurt Miller and
              Wouter Boomsma and
              Bradley Dice and
              Kostiantyn Lapchevskyi and
              Maurice Weiler and
              Michał Tyszkiewicz and
              Simon Batzner and
              Dylan Madisetti and
              Martin Uhrin and
              Jes Frellsen and
              Nuri Jung and
              Sophia Sanborn and
              Mingjian Wen and
              Josh Rackers and
              Marcel Rød and
              Michael Bailey},
    title = {Euclidean neural networks: e3nn},
    month = apr,
    year = 2022,
    publisher = {Zenodo},
    version = {0.5.0},
    doi = {10.5281/zenodo.6459381},
    url = {https://doi.org/10.5281/zenodo.6459381}
}

@article{e3x,
  title={\texttt{E3x}: $\mathrm{E}(3)$-Equivariant Deep Learning Made Easy},
  author={Unke, Oliver T. and Maennel, Hartmut},
  journal={arXiv preprint arXiv:2401.07595},
  year={2024}
}

@misc{Luo-24,
      title={Enabling Efficient Equivariant Operations in the Fourier Basis via Gaunt Tensor Products}, 
      author={Shengjie Luo and Tianlang Chen and Aditi S. Krishnapriyan},
      year={2024},
      eprint={2401.10216},
      archivePrefix={arXiv},
      primaryClass={cs.LG},
      url={https://arxiv.org/abs/2401.10216}, 
}

@inproceedings{passaro2023escn,
author = {Passaro, Saro and Zitnick, C. Lawrence},
title = {{Reducing SO(3) Convolutions to SO(2) for Efficient Equivariant GNNs}},
year = {2023},
publisher = {JMLR.org},
booktitle = {Proceedings of the 40th International Conference on Machine Learning},
articleno = {1140},
numpages = {19},
location = {Honolulu, Hawaii, USA},
series = {ICML'23}
}

@inproceedings{
xie2025the,
title={The Price of Freedom: Exploring Expressivity and Runtime Tradeoffs in Equivariant Tensor Products},
author={YuQing Xie and Ameya Daigavane and Mit Kotak and Tess Smidt},
booktitle={Forty-second International Conference on Machine Learning},
year={2025},
url={https://openreview.net/forum?id=EvIwwGYTLc}
}

@misc{xie2026asymptoticallyfastclebschgordantensor,
      title={Asymptotically Fast Clebsch-Gordan Tensor Products with Vector Spherical Harmonics}, 
      author={YuQing Xie and Ameya Daigavane and Mit Kotak and Tess Smidt},
      year={2026},
      eprint={2602.21466},
      archivePrefix={arXiv},
      primaryClass={cs.LG},
      url={https://arxiv.org/abs/2602.21466}, 
}

@misc{heyraud2026integralformulasvectorspherical,
      title={Integral Formulas for Vector Spherical Tensor Products}, 
      author={Valentin Heyraud and Zachary Weller-Davies and Jules Tilly},
      year={2026},
      eprint={2603.08630},
      archivePrefix={arXiv},
      primaryClass={cs.LG},
      url={https://arxiv.org/abs/2603.08630}, 
}

@misc{bochkarev2026fastcontractedclebschgordantensor,
      title={Fast contracted Clebsch--Gordan tensor products for equivariant graph neural networks}, 
      author={Anton Bochkarev and Yury Lysogorskiy and Ralf Drautz},
      year={2026},
      eprint={2605.15073},
      archivePrefix={arXiv},
      primaryClass={physics.comp-ph},
      url={https://arxiv.org/abs/2605.15073}, 
}

@inbook{Womersley2018,
  title = {Efficient Spherical Designs with Good Geometric Properties},
  ISBN = {9783319724560},
  url = {http://dx.doi.org/10.1007/978-3-319-72456-0_57},
  DOI = {10.1007/978-3-319-72456-0_57},
  booktitle = {Contemporary Computational Mathematics - A Celebration of the 80th Birthday of Ian Sloan},
  publisher = {Springer International Publishing},
  author = {Womersley,  Robert S.},
  year = {2018},
  pages = {1243–1285}
}

@misc{fu2025learningsmoothexpressiveinteratomic,
      title={Learning Smooth and Expressive Interatomic Potentials for Physical Property Prediction}, 
      author={Xiang Fu and Brandon M. Wood and Luis Barroso-Luque and Daniel S. Levine and Meng Gao and Misko Dzamba and C. Lawrence Zitnick},
      year={2025},
      eprint={2502.12147},
      archivePrefix={arXiv},
      primaryClass={physics.comp-ph},
      url={https://arxiv.org/abs/2502.12147}, 
}

@misc{wood2026umafamilyuniversalmodels,
      title={UMA: A Family of Universal Models for Atoms}, 
      author={Brandon M. Wood and Misko Dzamba and Xiang Fu and Meng Gao and Muhammed Shuaibi and Luis Barroso-Luque and Kareem Abdelmaqsoud and Vahe Gharakhanyan and John R. Kitchin and Daniel S. Levine and Kyle Michel and Anuroop Sriram and Taco Cohen and Abhishek Das and Ammar Rizvi and Sushree Jagriti Sahoo and Zachary W. Ulissi and C. Lawrence Zitnick},
      year={2026},
      eprint={2506.23971},
      archivePrefix={arXiv},
      primaryClass={cs.LG},
      url={https://arxiv.org/abs/2506.23971}, 
}

@inproceedings{li2026eformer,
  title={E2Former: An Efficient and Equivariant Transformer with Linear-Scaling Tensor Products},
  author={Yunyang Li and Lin Huang and Zhihao Ding and Xinran Wei and Chu Wang and Han Yang and Zun Wang and Chang Liu and Yu Shi and Peiran Jin and Tao Qin and Mark Gerstein and Jia Zhang},
  booktitle={The Thirty-ninth Annual Conference on Neural Information Processing Systems},
  year={2025},
  url={https://openreview.net/forum?id=ls5L4IMEwt}
}

@misc{huang2026e2formerv2ontheflyequivariantattention,
      title={E2Former-V2: On-the-Fly Equivariant Attention with Linear Activation Memory}, 
      author={Lin Huang and Chengxiang Huang and Ziang Wang and Yiyue Du and Chu Wang and Haocheng Lu and Yunyang Li and Xiaoli Liu and Arthur Jiang and Jia Zhang},
      year={2026},
      eprint={2601.16622},
      archivePrefix={arXiv},
      primaryClass={cs.LG},
      url={https://arxiv.org/abs/2601.16622}, 
}

@misc{qu2026recipescalableattentionbasedmlips,
      title={A recipe for scalable attention-based MLIPs: unlocking long-range accuracy with all-to-all node attention}, 
      author={Eric Qu and Brandon M. Wood and Aditi S. Krishnapriyan and Zachary W. Ulissi},
      year={2026},
      eprint={2603.06567},
      archivePrefix={arXiv},
      primaryClass={cs.LG},
      url={https://arxiv.org/abs/2603.06567}, 
}

@misc{elhag2026learninginteratomicpotentialsexplicit,
      title={Learning Inter-Atomic Potentials without Explicit Equivariance}, 
      author={Ahmed A. Elhag and Arun Raja and Alex Morehead and Samuel M. Blau and Hongtao Zhao and Christian Tyrchan and Eva Nittinger and Garrett M. Morris and Michael M. Bronstein},
      year={2026},
      eprint={2510.00027},
      archivePrefix={arXiv},
      primaryClass={cs.LG},
      url={https://arxiv.org/abs/2510.00027}, 
}

@misc{Shankar23,
      title={Importance of equivariant and invariant symmetries for fluid flow modeling}, 
      author={Varun Shankar and Shivam Barwey and Zico Kolter and Romit Maulik and Venkatasubramanian Viswanathan},
      year={2023},
      eprint={2307.05486},
      archivePrefix={arXiv},
      primaryClass={physics.flu-dyn},
      url={https://arxiv.org/abs/2307.05486}, 
}

@misc{Maennel2024,
      title={Complete and Efficient Covariants for 3D Point Configurations with Application to Learning Molecular Quantum Properties}, 
      author={Hartmut Maennel and Oliver T. Unke and Klaus-Robert Müller},
      year={2024},
      eprint={2409.02730},
      archivePrefix={arXiv},
      primaryClass={cs.LG},
      url={https://arxiv.org/abs/2409.02730}, 
}

@misc{jraph2020github,
  author = {Jonathan Godwin and Thomas Keck and Peter Battaglia and Victor Bapst and Thomas Kipf and Yujia Li and Kimberly Stachenfeld and Petar Veli\v{c}kovi\'{c} and Alvaro Sanchez-Gonzalez},
  title = {{J}raph: {A} library for graph neural networks in jax.},
  url = {http://github.com/deepmind/jraph},
  version = {0.0.1.dev},
  year = {2020},
}

@article{TPU-Jouppi17,
  title = {In-Datacenter Performance Analysis of a Tensor Processing Unit},
  volume = {45},
  ISSN = {0163-5964},
  url = {http://dx.doi.org/10.1145/3140659.3080246},
  DOI = {10.1145/3140659.3080246},
  number = {2},
  journal = {ACM SIGARCH Computer Architecture News},
  publisher = {Association for Computing Machinery (ACM)},
  author = {Jouppi,  Norman P. and Young,  Cliff and Patil,  Nishant and Patterson,  David and Agrawal,  Gaurav and Bajwa,  Raminder and Bates,  Sarah and Bhatia,  Suresh and Boden,  Nan and Borchers,  Al and Boyle,  Rick and Cantin,  Pierre-luc and Chao,  Clifford and Clark,  Chris and Coriell,  Jeremy and Daley,  Mike and Dau,  Matt and Dean,  Jeffrey and Gelb,  Ben and Ghaemmaghami,  Tara Vazir and Gottipati,  Rajendra and Gulland,  William and Hagmann,  Robert and Ho,  C. Richard and Hogberg,  Doug and Hu,  John and Hundt,  Robert and Hurt,  Dan and Ibarz,  Julian and Jaffey,  Aaron and Jaworski,  Alek and Kaplan,  Alexander and Khaitan,  Harshit and Killebrew,  Daniel and Koch,  Andy and Kumar,  Naveen and Lacy,  Steve and Laudon,  James and Law,  James and Le,  Diemthu and Leary,  Chris and Liu,  Zhuyuan and Lucke,  Kyle and Lundin,  Alan and MacKean,  Gordon and Maggiore,  Adriana and Mahony,  Maire and Miller,  Kieran and Nagarajan,  Rahul and Narayanaswami,  Ravi and Ni,  Ray and Nix,  Kathy and Norrie,  Thomas and Omernick,  Mark and Penukonda,  Narayana and Phelps,  Andy and Ross,  Jonathan and Ross,  Matt and Salek,  Amir and Samadiani,  Emad and Severn,  Chris and Sizikov,  Gregory and Snelham,  Matthew and Souter,  Jed and Steinberg,  Dan and Swing,  Andy and Tan,  Mercedes and Thorson,  Gregory and Tian,  Bo and Toma,  Horia and Tuttle,  Erick and Vasudevan,  Vijay and Walter,  Richard and Wang,  Walter and Wilcox,  Eric and Yoon,  Doe Hyun},
  year = {2017},
  pages = {1–12}
}

@article{Lysogorskiy2026,
  title = {Graph atomic cluster expansion for foundational machine learning interatomic potentials},
  volume = {12},
  ISSN = {2057-3960},
  url = {http://dx.doi.org/10.1038/s41524-026-01979-1},
  DOI = {10.1038/s41524-026-01979-1},
  number = {1},
  journal = {npj Computational Materials},
  publisher = {Springer Science and Business Media LLC},
  author = {Lysogorskiy,  Yury and Bochkarev,  Anton and Drautz,  Ralf},
  year = {2026},
  month = Feb 
}

@misc{liao2026equiformerv3scalingefficientexpressive,
      title={EquiformerV3: Scaling Efficient, Expressive, and General SE(3)-Equivariant Graph Attention Transformers}, 
      author={Yi-Lun Liao and Alexander J. Hoffman and Sabrina C. Shen and Alexandre Duval and Sam Walton Norwood and Tess Smidt},
      year={2026},
      eprint={2604.09130},
      archivePrefix={arXiv},
      primaryClass={cs.LG},
      url={https://arxiv.org/abs/2604.09130}, 
}

%%%%%%%%%%%%%%%%%%%%%%%%%%%%%%%%%%%%%%%%%%%%%%%%%%%%%%%%%%%%

\appendix

\section{Mathematical Background} 
In this appendix, we provide further theoretical details on the 
core components of {\tt e3j}. Representation theory of the rotation and Euclid groups had ground-breaking applications in Quantum Mechanics, where they 
notably provided a first derivation of the Hydrogen energy levels and their degeneracy, see e.g. Wigner \citep{Wigner-59}. 
For more contemporary introductions 
to the subject, readers may refer to 
\citep{Hall-2000, e3x}.

\subsection{Euclidean equivariance}

The Euclid group $E_3 = O_3 \ltimes \R^3$ describes the possible changes of frames of reference over the Euclidean 3-space, i.e. the compositions of translations, rotations and reflections. 
%$E_3$ naturally acts on the {\em orthonormal frame bundle} over $\R^3$, which consists of orthonormal tangent frames (transforming under the orthogonal group $O_3$) above base points of $\R^3$. Once an origin and reference frame have been chosen, the orthonormal frame bundle actually identifies with the Euclid group.
Euclidean equivariance is the property by which a function on the Euclidean space transforms consistently with its inputs upon any {\em Euclidean transform} $g \in E(3)$, for instance, with a force function ${\bf  F}({\bf r}, {\bf z})$: 
\begin{equation} \label{eq-forces}
{\bf F}(g \cdot {\bf r}, {\bf z}) = g \cdot {\bf F}({\bf r}, {\bf z}).
\end{equation}
where ${\bf r} \in \R^{n\times 3}$ denotes a matrix of atomic positions, ${\bf z} \in \N^n$ denotes a vector of atomic numbers, and $n$ is the number of atoms of the system or region of interest. 

In general, the Euclidean group may not only act on ($n$ copies of) $\R^3$, but also on general (real or complex) vector spaces called {\em representations} of $E_3$ (also called $E_3$-modules): they consist of pairs $(V, \rho)$ where the vector space $V$ is equipped with a smooth group morphism $\rho$ mapping any Euclidean transform $g \in E_3$ to an invertible matrix $\rho_g \in GL(V)$.
A function ${\bf F} : V \to V'$ is called {\em equivariant} if the following diagram is commutative:
\begin{equation}
\begin{tikzcd}
V \rar{{\bf F}} \dar[swap]{\rho_g} & V' \dar{\rho'_g} \\
V \rar{{\bf F}} & V' 
\end{tikzcd}
\end{equation}
While morphisms of $E_3$-representations are usually assumed linear, note that the above definition of equivariance applies 
to non-linear functionals just as well, in particular polynomial 
functionals which are of particular importance in 
the classification of $E_3$-representations.

A fundamental result is that any orthogonal representation 
$V$ of $O_3$ can be decomposed into a direct sum of irreducible representations or {\em irreps}, each {\em irrep} being chosen among a well known classification of possible fundamental types 
(related to harmonic polynomials over $\R^3$). 
Note that the dimensions of irreducible representations 
translate into tangible observables in quantum chemistry (QC), 
where $l$ and $2l + 1$ respectively determine the symmetry (S, P, D...) and degeneracy (half the maximal number of occupying electrons) 
of an energy level in a hydrogen-like atom.

\subsection{Harmonic polynomials} 
\label{sec:theory:harmonics}

\paragraph{Definition.}
The representation theory of $SO_3$ is closely related to the 
harmonic polynomials of $\C[x, y, z]$. 
Homogeneous, degree-$l$ polynomials are finite-dimensional 
vector spaces, naturally equipped with an $SO_3$ action. Because the 
Laplacian operator $\Delta$ (trace of the hessian) is also invariant
under $SO_3$, the space $\Y_l \subset \C_l[x, y, z]$ of degree-l 
harmonic polynomials (satisfying $\Delta P= 0)$ 
is a sub-representation, i.e. a subspace stable under $SO_3$.

$$ \Delta Y_l^m = 
\frac{\partial^2 Y_l^m}{\partial x^2}
+ \frac{\partial^2 Y_l^m}{\partial y^2}
+ \frac{\partial^2 Y_l^m}{\partial z^2} = 0
$$

Any irreducible representation of $SO_3$ (i.e. a "smallest" vector space with an $SO_3$-action, having no other stable strict subspace than 0) is 
isomorphic to some $\Y_l$, of odd dimension $2l + 1$. Any larger 
representation of $SO_3$ can be decomposed as a direct sum of irreducible representations $\bigoplus_l (\Y_l)^{k_l}$.

\paragraph{Equivariance.} 

For every degree $l$, the maps $\mathbf{Y}_l : \R^3 \to \mathbb{Y}_l \simeq \C^{2l+1}$ 
are {\em equivariant non-linear embeddings}, 
typically used as a basis for learning more complex non-linear equivariant 
representations $f_\theta : V \to V'$ within e.g. deep MLIP networks. 
This means that for every rotation $g \in SO_3$, one may construct the so-called 
{\em Wigner D-matrix} $D_g$, acting 
on the $2l+1$ space of polynomial activations ${\bf Y}_l$ in 
the following commutative diagram: 
\begin{equation}
    \begin{tikzcd}
    \R^3 \dar[swap]{g} \rar{{\bf Y}_l} & \Y_l \dar{D_g} \\
    \R^3 \rar{{\bf Y}_l} & \Y_l \\
    \end{tikzcd}
\end{equation}
The spaces $\mathbb{Y}_l$ consist of the basic pieces of any Euclidean representation 
$V$, as any irreducible representation of the group of rotations $O_3$ is isomorphic 
to some $\mathbb{Y}_l$ for some $l$. 
Additionally, irreducible $E_3$ representations carry a parity label $\pm$ (even/odd) 
dictating whether reflections act with a sign or as the identity. 
See Appendix \ref{apx:reconstructions} for further mathematical details.

\subsection{Key Operations in Scope}\label{apx:operations-scope}

Equivariant architectures typically enforce the equivariance
constraint ~\eqref{eq-forces} by a few common design patterns
and building blocks, the {\tt e3j}  package provides a harmonized and functional API around 
these:  
\begin{itemize} 
\item \textbf{Harmonics:} Restrict available geometric information to the edge vectors 
$({\bf r}_{ab}) \in \R^{n_E\times3}$. This already 
enforces translation invariance. Then expand edge vectors
${\bf r}_{ab} \in \R^3$ with {\em harmonic embeddings}
${\bf Y}_l({\bf r}_{ab})$ where ${\bf Y}_l$ denotes one bank of $2l+1$
rotation-equivariant activation filters built from the $2l+1$ degree-$l$ harmonic polynomials,  typically concatenated over degrees 
$l = 0, \dots, l_{\rm max}$ with $l_{\rm max}$ rarely exceeding 3:
\begin{equation}
{\bf Y}_l({\bf r}) = \big[\, Y_{lm}({\bf r}) \,|\, m=-l \dots l\,\big]
\end{equation} 

\item \textbf{Linear mixing:}
Rescale or mix channels and multiplicities linearly 
between irreducible features of a same degree $l$. 
While rarely a bottleneck by itself, there is opportunity for scalar 
mixings to be fused e.g. within message-passing operations, where 
edge scalars representing chemical species and radial embeddings 
of interatomic distances are coupled with rotation-equivariant features.
\begin{equation}
    {\tt L}_W({\bf x})^{lm,k} = \sum_{k'm'} W^{km}_{k'm'} {\bf x}^{lm',k'}
\end{equation}

\item \textbf{Tensor product: }Couple latent equivariant features with {\em Clebsch-Gordan tensor products} ${\bf z} = {\bf x} \otimes_C {\bf y}$, where 
${\bf x}$ may for instance denote latent node features from the 
previous layer and ${\bf y}$ the harmonic embedding of an edge 
vector, or a more general latent feature vectors organized in irreps. When ${\bf x}$ and ${\bf y}$ are irreducibles of degree 
$l$ and $l'$, their tensor product ${\bf z}$ is obtained from the 
$2 \min(l, l') + 1$ bilinear pairings (or "paths") of output 
degrees $L = |l-l'|, \dots, l+l'$, given by:
\begin{equation}
    ({\bf x} \otimes_C {\bf y})^{LM} = \sum_{m+m'=M} C^{LM}_{lm,l'm'} \, {\bf x}^{lm} \,{\bf y}^{l'm'}
\end{equation}
where $C^{LM}_{lm,l'm'}$ are the Clebsch-Gordan coefficients.

\item \textbf{Message passing:} Aggregate the edge-wise features, usually computed through combination of harmonics projection, tensor product and linear or scalar mixing. The message passing aggregation tends to be the operational bottleneck once efficient tensor product operations are implemented. A typical message passing layer in the context of equivariant GNNs can be written as:
\begin{equation} \label{eq:message-passing-linear}
    {\bf m{\prime}}_b = \sum_{a \in \mathcal{N}(b)}  \, {\tt L}_{{\bf s}_{ab}} 
    \left({\bf x}_a\otimes_C {\bf Y}({\bf r}_{ab})\right) 
\end{equation}
where ${\bf Y}({\bf r}_{ab})= \bigoplus_{l}{\bf Y}_l({\bf r}_{ab})$ and ${\tt L}_{{\bf s}_{ab}}$ denotes a 
linear mixing with radial edge scalars ${\bf s}_{ab}$, which typically depend on 
interatomic distances $||{\bf r}_{ab}||$ 
via a radial basis function (RBF) embedding followed by a multi-layer perceptron (MLP).
\end{itemize}

While we have used MLIP as a testing ground for realistic workloads, we have not yet included operations specifically targeted at particular MLIP models. The main example is the Symmetric Contraction module in MACE. We did however include benchmarks with the existing {\tt cuEquivariance} Symmetric Contraction kernel in appendix \ref{apx:symmetric-contraction}.

\section{Comparison with Cubically Scaling Tensor Products} \label{apx:cubically-scaling}

\subsection{Gaunt and Vector Signal Tensor Products} \label{apx:gaunt-comparison}

In this section, we benchmark the {\tt e3j} tensor product against the Gaunt tensor product (GTP) \citep{Luo-24} and the vector signal tensor product (VSTP) \citep{xie2026asymptoticallyfastclebschgordantensor, heyraud2026integralformulasvectorspherical,bochkarev2026fastcontractedclebschgordantensor}, which evaluate the tensor product as an integral over the sphere rather than a contraction over Clebsch-Gordan coefficients. The GTP and VSTP are complementary: the GTP covers only the symmetric paths, where the sum $l_1 + l_2 + l_3$ of all of the irreps entering into the tensor product is even, while the VSTP covers only the skew-symmetric paths where the sum is odd.

\begin{figure}[h]
    \centering
    \includegraphics[width=0.95\linewidth]{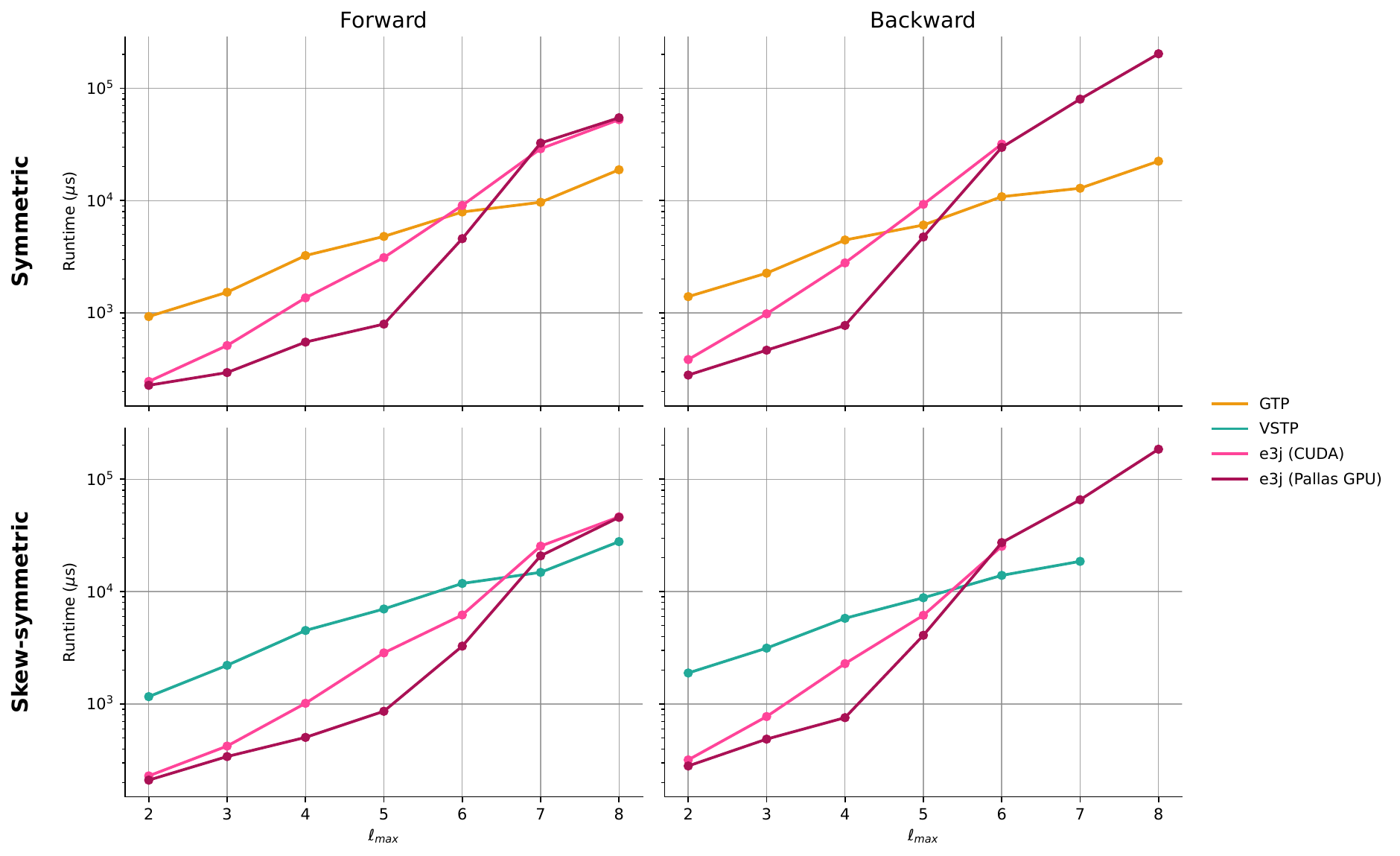}
    \caption{
    \textbf{Comparison with Gaunt and Vector Signal Tensor Products.} Benchmarks show the runtime scaling with $L= l_{max}$ with a fixed batch size $B = 32{,}768$ and channels $C=128$. Every Clebsch-Gordan path $(l_1, l_2, l_3)$ is symmetric or skew-symmetric, according to the parity of $l_1 + l_2 + l_3$. The symmetric panel, where the sum is even, benchmarks the {\tt e3j} tensor product on symmetric paths against the Gaunt tensor product (GTP) \citep{Luo-24}, whilst the skew-symmetric panel benchmarks against the vector signal tensor product (VSTP) \citep{xie2026asymptoticallyfastclebschgordantensor,heyraud2026integralformulasvectorspherical}.  Both the GTP and the VSTP are evaluated through integral formulas on a spherical $t$-design quadrature~\citep{Womersley2018}. Runtimes are obtained on a single NVIDIA H100.} 
    \label{fig:gaunt-comparison}
\end{figure}

Efficient implementations of both the VSTP and GTP rely on the tensor product emitting a single copy of each output degree rather than one per path, and, relatedly, that a weighted tensor product has weights taking a factorised form $w^{l_3}_{l_1 l_2} = a_{l_1} b_{l_2} c_{l_3}$. In this benchmark, we therefore collapse the ${\tt e3j}$ output by degree to ensure the output spaces match in all cases. We also choose to time the tensor product, since factorised weights act as linear maps on the input and output spaces that are identical in all three cases. In practice, we compute the GTP and VSTP from their integral formulas,  which are evaluated on a spherical $t$-design quadrature \citep{Womersley2018}. 

To match the paths of the GTP and VSTP, we utilize the parity selection rule $p_1 p_2 = p_3$ that the Clebsch-Gordan product already enforces. With natural-parity irreps $p_l= (-1)^l$ for both the inputs and the targets, the parity rule enforces $(-1)^{l_1 + l_2 + l_3} =1$ and keeps exactly the symmetric paths. Conversely, with the anti-natural $p_l = (-1)^{l+1}$ irreps, only the skew-symmetric paths are allowed. 

The results are shown in Figure \ref{fig:gaunt-comparison}. For $L\leq 3$, where the cost of the quadrature operations dominates, the {\tt e3j} kernel is $2$ to $4$ times faster than the GTP and $4$ to $6$ times faster than the VSTP. As $L$ grows, the theoretical scaling in $L$ becomes relevant ($O(L^4)$ for quadrature methods vs $O(L^5)$ for sparse Clebsch-Gordan contraction). The forward pass curves cross at $L=6$ and $L=7$ for the GTP/VSTP respectively, and at $L=5$ and $L=6$ for the backward, showing that {\tt e3j} remains faster in the regime practical for MLIP applications. It is worth noting that there are also other implementations of the GTP and VSTP that have better theoretical scaling in $L$ \citep{xie2026asymptoticallyfastclebschgordantensor}, but that we find to be slower in practice over this range of degrees. 

\subsection{Matrix Tensor Products from {\tt e3x}} \label{apx:e3x-comparison}

Another tensor product operation with an efficient $O(L^4)$ scaling has been proposed by the authors of the {\tt e3x} library~\citep{Maennel2024, e3x}. We follow Xie et al.~\citep{xie2025the} and refer to this operation as Matrix Tensor Product (MTP). The MTP is implemented in the {\tt FusedTensor} module of {\tt e3x}. This operation computes the tensor product between feature vectors in three steps. First, considering the operation couples two input vectors containing irreps features of order $0\leq l \leq L$, each input vector is mapped to a $(2\tilde l + 1)\times (2\tilde l + 1)$ square matrix, with $\tilde l = \left\lceil L / 2 \right \rceil$. This map corresponds to the isomorphism 
\begin{equation}
\bigoplus_{l=0}^{L} \mathbb{Y}_{l} \cong \mathbb{Y}_{\tilde{l}} \otimes \mathbb{Y}_{\tilde{l}},
\end{equation}
where elements of the right hand-side space can be seen as square matrices produced by an outer product of two feature vectors in $\mathbb{Y}_{\tilde{l}}$. Then, the square matrices encoding the inputs are multiplied. Finally, the resulting square matrix is mapped back to the output feature vector. The matrix product exhibits an efficient $O(L^3)$ scaling, but the conversions between square matrices and feature vectors yield an overall scaling of $O(L^4)$. 

In this section, we benchmark the {\tt e3j} tensor product against the MTP. Note that for a given irrep path, the MTP is proportional to the GTP-VSTP operations, although the proportionality coefficient sometimes vanishes, so that the MTP is in general strictly less expressive than the GTP-VSTP. We compare the FusedTensor module of {\tt e3j} against the same {\tt e3j} tensor product used in the benchmark of Appendix~\ref{apx:gaunt-comparison}. As the {\tt FusedTensor} implementation includes the path weights $w^{l_3}_{l_1 l_2}$, we include linear mixings on both the inputs and the output feature vectors of the {\tt e3j} tensor product, so that both operations are strictly equivalent. Note that these additional linear mixings account for its higher runtime compared to the version of Appendix~\ref{apx:gaunt-comparison}.

\begin{figure}[h]
    \centering
     \includegraphics[width=\linewidth]{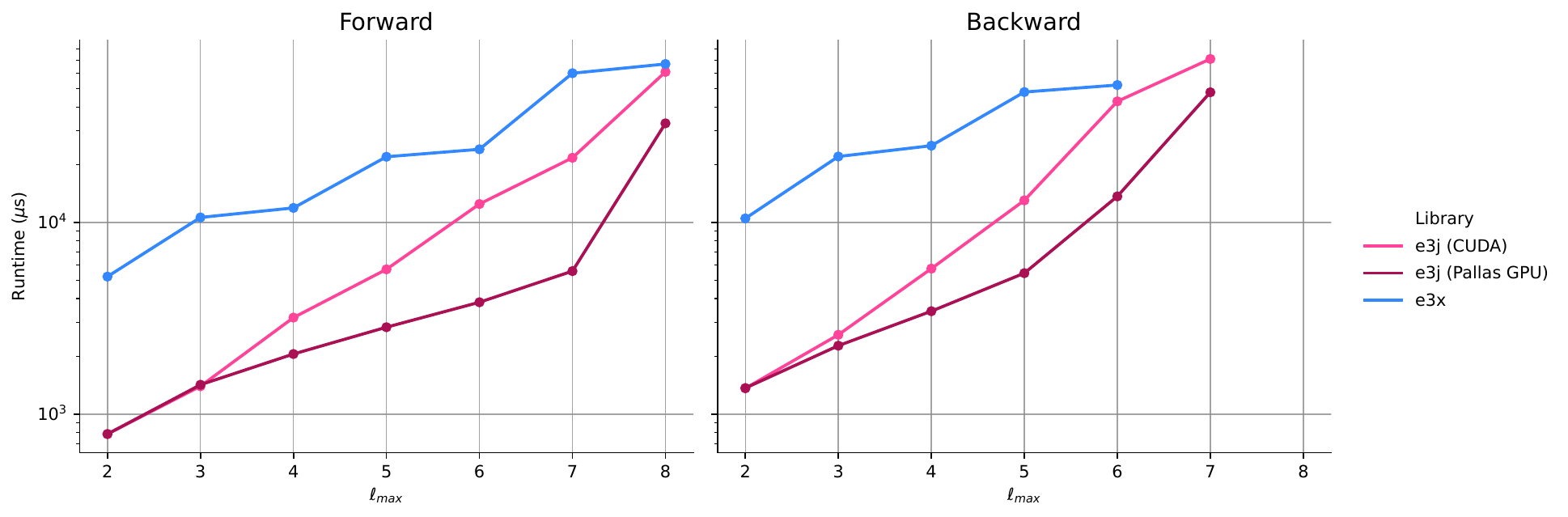}

    \caption{
    \textbf{Comparison of {\tt e3j} and {\tt e3x}'s {\tt FusedTensor} tensor products.} Benchmarks show the runtime scaling with $L= l_{max}$ with a fixed batch size $B = 32{,}768$ and channels $C=128$. Every Clebsch-Gordan path $(l_1, l_2, l_3)$ is symmetric. Runtimes are obtained on a single NVIDIA H100.} 
    \label{fig:e3x-comparison}
\end{figure}

The results are shown in Figure~\ref{fig:e3x-comparison}. For $L\leq 8$ the {\tt e3j} tensor product is faster than the {\tt e3x} implementation. As $L$ grows, the efficient $O(L^4)$ scaling of the MTP reduces the gap with the {\tt e3j} implementation, though {\tt e3j} remains faster on this domain.

\subsection{SO2 Convolution} \label{apx:so2-comparison}

In order to compare with SO2 convolution fairly we evaluated the {\tt e3j} 
convolution implementations with a set of coefficients that collapses output multiplicities, 
i.e. sums all isomorphic 
copies of a given irreducible output space together. 
The reduced output feature dimension means a cost on expressivity, at the benefit of 
a smaller memory traffic.

\begin{figure}[ht]
    \centering
    \includegraphics[width=\linewidth]{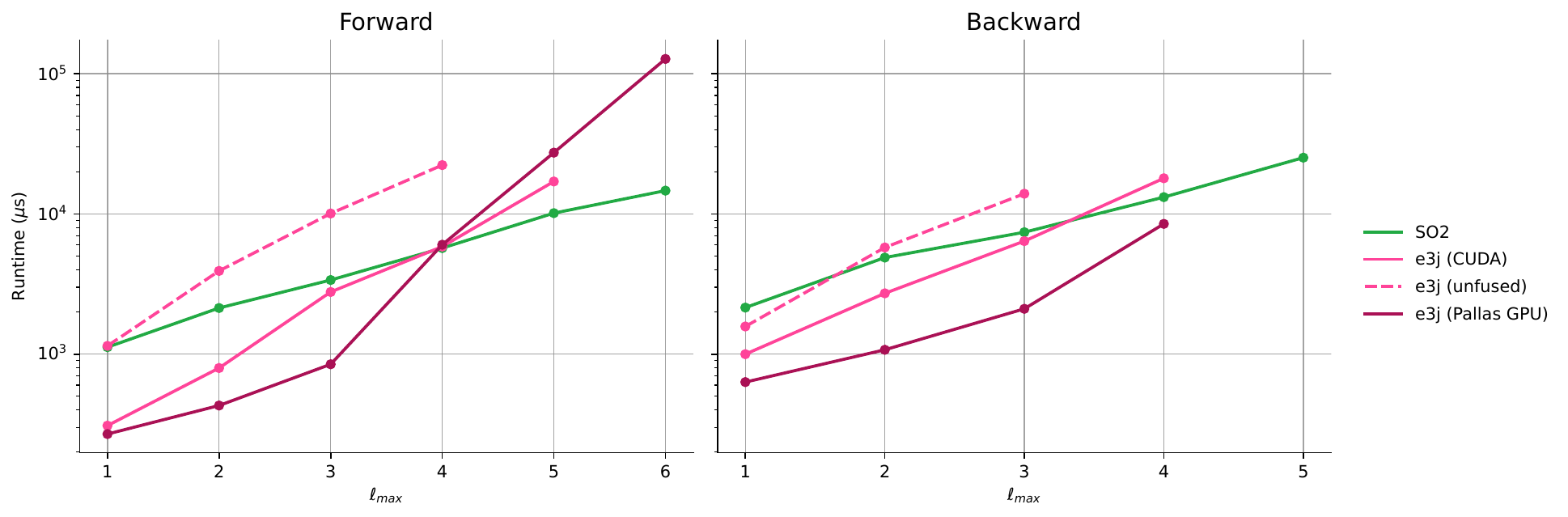}
    \caption{\textbf{Comparison of SO2 convolution with Clebsch-Gordan convolution.}
        Benchmarks show the runtime scaling with $L = l_{max}$ with fixed number of nodes 
        $N=2048$, number of edges $N_e = 45  \times N$, and fixed number of channels $C=128$.
        In contrast to other convolution benchmarks with third-party baselines, output multiplicities 
        are collapsed to one by providing an ad-hoc set of sparse COO coefficients to match a JAX implementation 
        of the SO2 convolution algorithm described in \citep{passaro2023escn}. Runtimes are obtained on a single NVIDIA H100.
    }
    \label{fig:so2-comparison}
\end{figure}

Note that while \texttt{e3j} convolution kernels natively support any set of coefficients, 
they have not been optimized for these smaller problem shapes. 
We however notice that \texttt{e3j} is between 4x and 6x faster at $\ell_{max} \leq 3$, 
while matching SO2 convolution at $\ell_{max} = 4$, see figure \ref{fig:so2-comparison}.
Although the XLA compilation of a plain JAX implementation of SO2 convolution already 
proves very efficient, 
a faithful comparison of the method would require similar low-level engineering on 
the SO2 convolution implementation.

\section{Additional End-to-end Benchmarks}\label{apx:additional-benchmarks}

\subsection{Symmetric contraction benchmarks} \label{apx:symmetric-contraction}

\begin{figure}[h]
    \centering
    \includegraphics[width=1\linewidth]{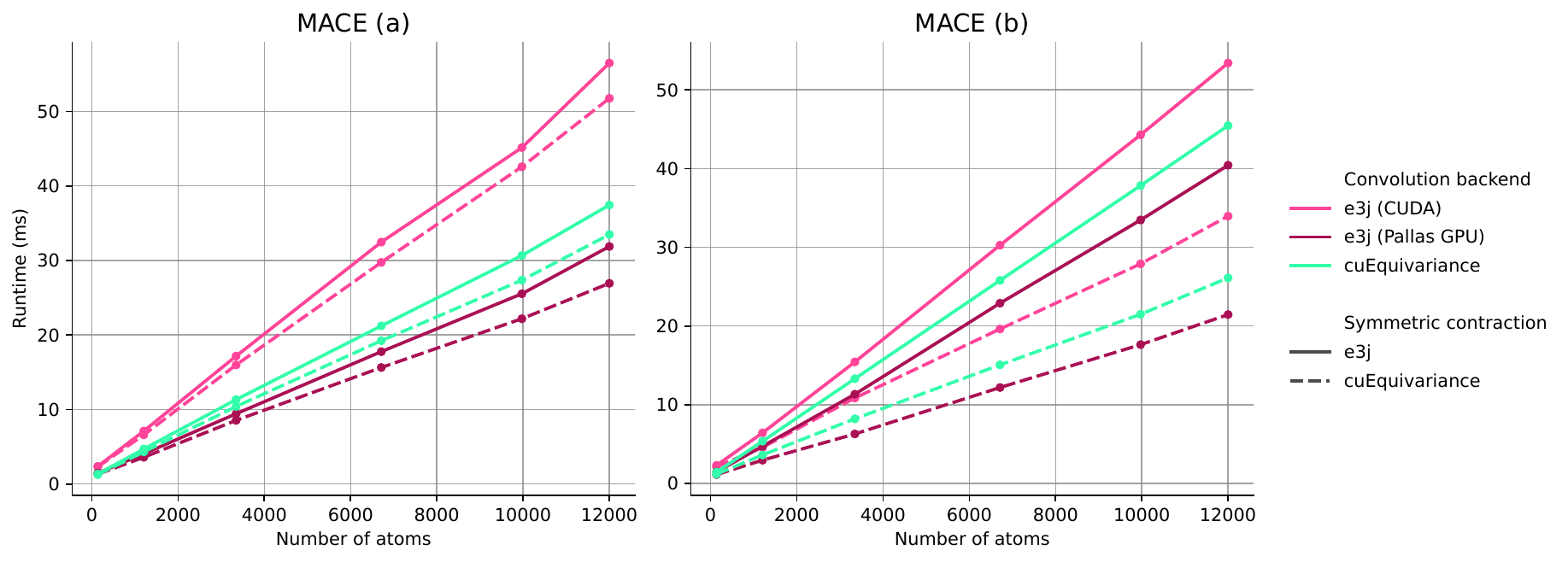}
    \caption{\textbf{Effect of the \texttt{SymmetricContraction} backend for the MACE model on GPU.}
    We compare the effect of replacing the naive {\tt e3j}-based power expansion, followed by 
    a species-wise linear projection of higher-order features, with the dedicated 
    {\tt SymmetricContraction} kernel of {\tt cuEquivariance}. The two variants (a) and 
    (b) of the MACE model are detailed in table \ref{table:hparams}. 
    }
    \label{fig:symmetric-contraction}
\end{figure}

Most of the MACE model benchmarks reported in the main text use the same implementation for 
the {\tt SymmetricContraction} operation, which consists of: 
\begin{itemize}
    \item a {\tt PowerExpansion} of node features (quadratic or cubic) using the tensor product 
    kernels of {\tt e3j} with channel mixing-mode {\tt MAP}\footnote{
        Note that not all CUDA backends expose this functionality, and that the batch and channel axes are not 
        contiguous in the optimal trailing channels layout.
    }, 
    \item a {\tt LinearIndexwise} projection of the concatenated higher-order features using 
    weights that depend on the atomic species.
\end{itemize}
While conceptually simple, this implementation is suboptimal as the power expansion generates 
large multiplicities that all collapse eventually after the linear projection steps. The 
GMEM materialization of the large intermediate array can be skipped by a dedicated  
{\tt SymmetricContraction} fusing both operations, as reported in figure \ref{fig:symmetric-contraction}. 

An optimized implementation may follow the algorithm proposed in the original MACE paper \citep{MACE}, 
in device code. The plain JAX implementation of this algorithm however performs worse than the naive power expansion 
and linear projection, once an efficient CGTP is available. The {\tt e3nn} force-inference benchmarks, 
illustrating the best efficiency one may reach with a pure JAX implementation, rely on this implementation. 
However, we cannot currently change the algorithm significantly without breaking numerical consistency. 
Restoring consistency requires a complex 1-to-1 transformation of learnable parameters and CG coefficient 
normalization choices to be resolved.

All NPT simulation benchmarks use the same {\tt e3j}-based {\tt SymmetricContraction}. 
While isolating the effect of the convolution backend, this allows us to load a single model 
checkpoint to run stable NPT simulations. 

At the time of writing, {\tt e3j} does not provide a dedicated {\tt SymmetricContraction} kernel.
Our current investigations seemed to show that inlining coefficients with a JIT compiled kernel (using Pallas or NVRTC)
seems necessary to reach the same performance as {\tt cuEquivariance}, given that our AOT compiled CUDA kernels 
reach due to the large number of coefficients and feature sizes that occur during this operation.

\subsection{Determinism and deviations of end-to-end predictions} \label{apx:determinism}

Non-deterministic message aggregation may prove very efficient, 
since it a allows a kernel to directly loop and distribute work over edges 
(instead of nesting a potentially imbalanced loop over neighbors inside a loop over receiver nodes) 
and the memory cache hierarchy may efficiently hide the latency of memory-locked atomic operations.
Valency imbalance for instance explains why the deterministic {\tt OpenEquivariance} kernel leads to slower simulations 
than the non-deterministic, when enforcing a static edge count with all padding edges joining a single padding node, 
despite being faster in non-padded unique benchmarks. 

In some downstream workflows (relaxation, geometry optimization, ...), 
deterministic predictions may however be a important requirement.
When relaxing a periodic cell (as in NPT simulations, see table \ref{table:npt_runtimes}) 
with a Monte-Carlo barostat, deterministic energy predictions are for instance a hard constraint: 
small energy fluctuations enter an exponential Boltzmann factor used for a Metropolis-Hastings rejection 
criterion, and non-deterministic message aggregation would lead to exploding simulations.

\begin{figure}[h]
    \centering
    \includegraphics[width=1\linewidth]{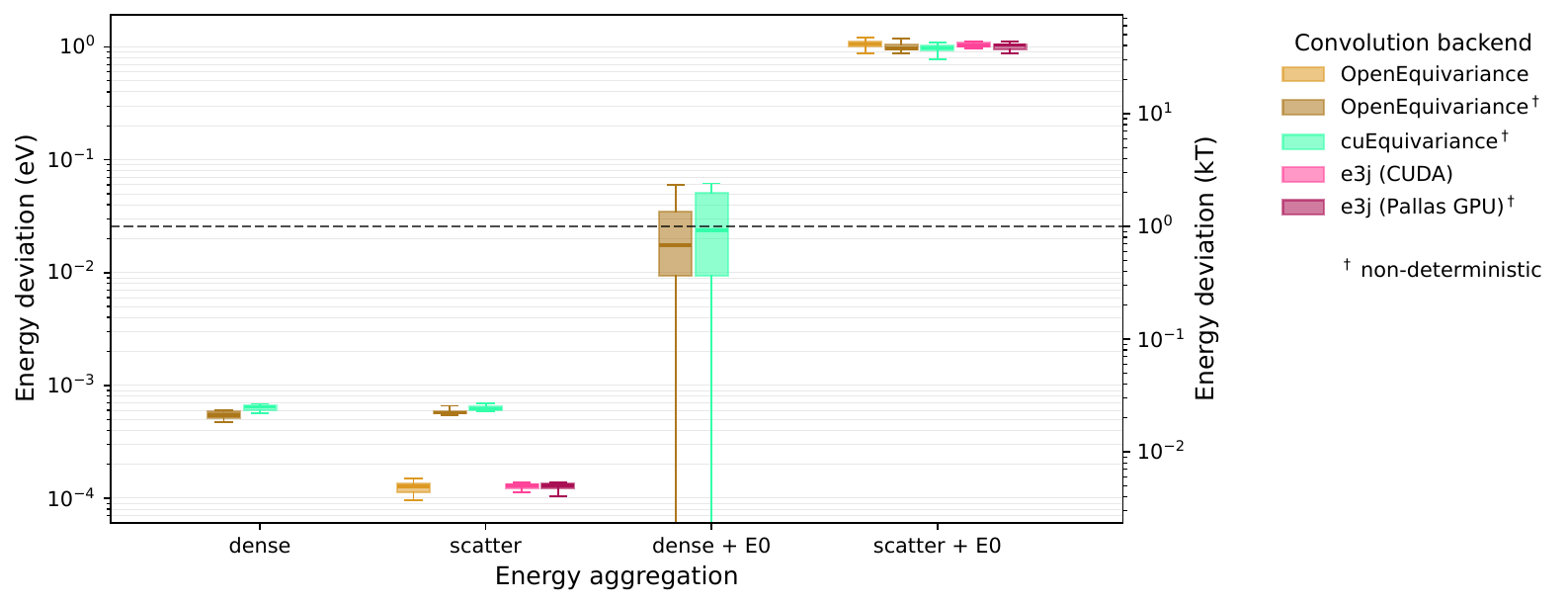}
    \caption{\textbf{Run-to-run energy deviation of a MACE model (a) on GPU.}
    The trained model used for NPT simulations (table \ref{table:npt_runtimes}) is evaluated 
    on a batch of 8 water boxes totaling 21,016 atoms and 889,840 edges, over a 100 times. 
    The experiment was repeated using different energy aggregation schemes: including or dropping 
    the constant atomic energies ($E_0$); using {\tt mlip}'s default scatter aggregation of node energies over graphs 
    or a dense matrix-vector product for the energy head.
    The plot represents 5th/95th percentiles as whiskers and 25th/75th percentiles as boxes.
    When a box is missing, it means that all runs were rigorously deterministic.
    Hyperparameters are detailed in table \ref{table:hparams}. 
    }
    \label{fig:energy_deviation}
\end{figure}

\begin{figure}[h]
    \centering
    \includegraphics[width=1\linewidth]{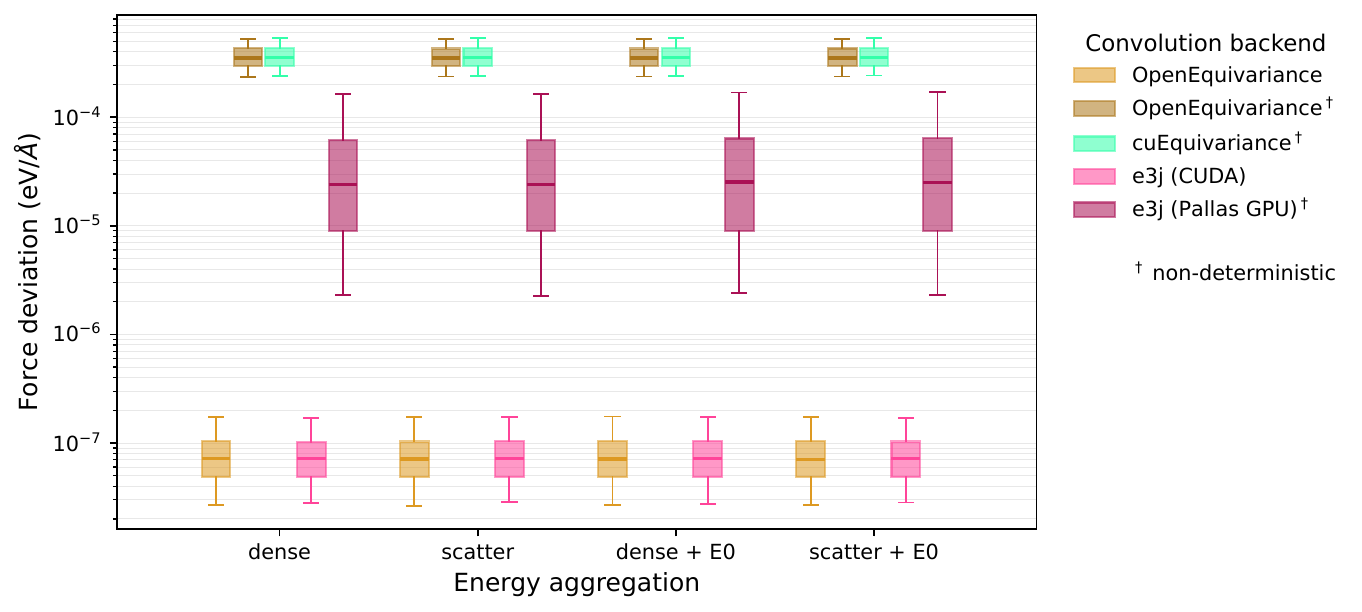}
    \caption{\textbf{Run-to-run force deviation of a MACE model (a) on GPU.}
    The trained model used for NPT simulations (table \ref{table:npt_runtimes}) is evaluated in 
    the same conditions as figure \ref{fig:energy_deviation}. Because XLA can eliminate the addition of atomic energies, 
    and because the VJP of the scatter operation is a deterministic gather operation, 
    we see that the addition of $E_0$ and the aggregation method have no effect on the deviation of forces.
    Even with deterministic convolution kernels, forces have non-zero deviations due to some forward 
    gather operations being transposed as non-deterministic scatter operations. 
    Interestingly the Pallas GPU kernel, only deterministic in the forward pass, lies in between. 
    }
    \label{fig:forces_deviation}
\end{figure}

In addition to the message-passing step, two sources of non-determinism or numerical noise may compound 
in the energy prediction: 
\begin{itemize}
    \item addition of atomic energies $E_0$, which are multiple orders of magnitude larger than the geometry-induced
    energy variations, and may truncate the constant relative precision of floating-points data types,
    \item aggregation of graph energies from node energy summands, which may scatter a very large number of contributions 
    to a few scalar numbers with a high degree of concurrency and collisions.
\end{itemize}
Both of these effects are analyzed in figure \ref{fig:energy_deviation}. 
In contrast, the force prediction may eliminate the addition of constants and replace non-deterministic scatter operations 
by deterministic gather operations in the computational graph, as illustrated by figure \ref{fig:forces_deviation}. 

We note that this mitigation of stochastic variations is not an automatic consequence of the differentiation process, 
since differentiating a mean-squared-error loss would presumably compound sources of non-determinism instead of 
eliminating most of them (through a product of the energy error with energy gradients). 
A rigorous analysis of the effect of non-determinism on model training behavior is however 
out of scope of the present work.

\section{Benchmarks details} \label{apx:benchmark_details}

\subsection{Throughput and HBM}
Our module-specific benchmarks mostly focus on kernel {\em throughput}, commonly defined as: 
\begin{equation}
    {\tt throughput} = 
    \frac{\tt sizeof(inputs) + sizeof(outputs)} {\tt runtime}
\end{equation}
In addition to being asymptotically independent of I/O size, 
throughput is also bounded by the so-called {\em global memory (GMEM) bandwidth}, 
corresponding to the ideal throughput of an optimal array copy. 
Global memory is the long-lived data bank used by 
the device processors to load/store I/O data. It is also called 
{\em high bandwidth memory} (HBM) in manufacturer specifications, given 
its practical importance in delivering the best I/O throughput possible.

The NVIDIA{\textregistered} H100 graphical processing units (GPUs), on which 
most of our experiments were performed, advertises about 
3.35 TB/s HBM. The Google\textregistered tensor processing units (TPUs) 
we could experiment with advertizes 1.20 TB/s HBM for v4 and 1.64 TB/s HBM for v6e (Trillium). 
Note constant technical progress is made on those characteristics,
and a fair comparison should compare devices from the same year, 
and weigh those metrics by affordability.

\subsection{Precision}

Although {\tt e3j} also provides {\tt float64} binaries, 
all benchmarks were carried with I/O arrays in single {\tt float32} precision, which proves enough to run stable molecular dynamics simulations. 
Note however that JAX may internally resort to half-precision arithmetic in tensor contractions ({\tt matmul}, {\tt einsum}, \dots) for faster execution, and caps to single-precision by default to avoid downsides incurred by undesired upcasts.

On normal random input, our experiments show that all equivariance backends, regardless of platform, lead to similar accuracies of order $4\times 10^{-8}$ and $1 \times 10^{-7}$ for tensor products and message-passing operations respectively, with respect to a common, deterministic {\tt float64} reference. 
The only exception is the {\tt e3nn} backend which only yields about $4 \times 10^{-4}$ elementwise accuracy when the environment variable ${\tt JAX\_DEFAULT\_MATMUL\_PRECISION}$ is not set to highest. 
This significant gap in precision should therefore be
considered when comparing ${\tt e3nn}$ with other backends in end-to-end benchmarks.

\subsection{Considerations regarding Pallas and CUDA kernels}
\label{sec:pallas-cuda}

To yield speedups on the Google\textregistered  tensor processing units (TPUs) which 
JAX targets as well, E3J also defines kernels written in Pallas, 
a domain-specific language (DSL) which 
is part of the JAX package and targets GPU and TPU compilation.
While TPU benchmarks can only compare Pallas kernels of E3J with E3NN \citep{e3nn_lib}, 
the GPU benchmarks may compare CUDA and Pallas implementations of E3J 
with other CUDA implementations such as NVIDIA CuEquivariance (TM) and 
OpenEquivariance \citep{openeq}. 
One advantage CUDA nonetheless brings 
over Pallas is the relative stability of compiled 
binaries and {\tt nvcc} toolchain over JAX version dependencies.

However, the Pallas language and Mosaic GPU compiler streamline the 
just-in-time (JIT) compilation of device code, enabling the production of very specialized 
and efficient kernels. In addition to problem shapes and sizes, that cannot be 
trivially defined from static parameters with ahead-of-time (AOT) compilation 
of a traditional CUDA kernel, JIT compilation from Python source lets one 
seamlessly {\em inline} the Clebsch-Gordan coefficients inside the produced 
assembly code. This can significantly reduce memory traffic from the unified 
SMEM/L1 cache.

Note that other solutions such as NVIDIA's NVRTC compiler enable JIT compilation 
of CUDA/\Cxx source, a solution leveraged by \texttt{OpenEquivariance}. 
Our CUDA kernels do not make use of JIT compilation at this time, and stream through 
coefficients as an actual array buffer loaded from global memory.

\subsection{MLIP integration}

Hyperparameters defining the models used in the end-to-end benchmarks are 
detailed in table \ref{table:hparams}, they corresponding to the configuration 
fields of the {\tt mlip} library 
\citep{MLIP-paper}.

While end-to-end MLIP integration gives the most significant results for applications,
it is also a complex task that needs to be carried carefully in order to preserve 
numerical predictions and faithfulness of comparisons. Furthermore, while MLIPs 
consist of a primary motivation for {\tt e3j}, we view the library as an all-purpose 
low-level tool whose usage may not be limited to the two particular architectures 
benchmarked in the present work.

All the MLIP models compared were checked to match numerically regardless of the convolution 
or tensor product backend. Comparison with third-party implementations of models, 
or comparisons substituting additional blocks of the MACE model (such as \texttt{SymmetricContraction}) has not been performed 
at this time, due to the significant effort required to gain solid confidence 
in the faithfulness of final comparisons, and its orthogonality
with the low-level engineering effort behind {\tt e3j}.
\begin{table}[h]
\caption{\bf Hyperparameters used in end-to-end MLIP benchmarks.}
\label{table:hparams}
\centering
\begin{minipage}{0.48\textwidth}
\centering
\begin{tabular}{ll}
\toprule
MACE (a)\\
\midrule
\texttt{num\_layers}    & 2   \\
\texttt{num\_channels}  & 128* \\
\texttt{correlation}    & \textbf{2}   \\
\texttt{node\_symmetry} & \textbf{2}   \\
\texttt{l\_max}         & 3   \\
\texttt{cutoff\_angstrom}        & 5              \\
\texttt{num\_rbf}                & 8              \\
\texttt{node\_gating}            & \texttt{true}  \\
\texttt{include\_pseudotensors}  & \texttt{false} \\
\bottomrule
\end{tabular}
\end{minipage}
\hfill
\begin{minipage}{0.48\textwidth}
\centering
\begin{tabular}{ll}
\toprule
MACE (b)\\
\midrule
\texttt{num\_layers}    & 2   \\
\texttt{num\_channels}  & 128* \\
\texttt{correlation}    & \textbf{3}   \\
\texttt{node\_symmetry} & \textbf{1}   \\
\texttt{l\_max}         & 3   \\
\texttt{cutoff\_angstrom}        & 5              \\
\texttt{num\_rbf}                & 8              \\
\texttt{node\_gating}            & \texttt{true}  \\
\texttt{include\_pseudotensors}  & \texttt{false} \\
\bottomrule
\end{tabular}
\end{minipage}
\centering
\vspace{1.5em}

\begin{tabular}{ll}
\toprule
NequIP \\
\midrule
\texttt{num\_layers}    & 5   \\
\texttt{num\_channels}  & 32* \\
\texttt{node\_irreps}   & {\smaller \texttt{2x0e + 2x0o + 1o + 1e + 2e + 2o}} \\
\texttt{l\_max}         & 2   \\
\texttt{cutoff\_angstrom}        & 5    \\
\texttt{num\_rbf}                & 8    \\
\bottomrule
\end{tabular}
\end{table}
\section{Tensor Product and Message Passing Convolution kernels} \label{apx:kernel_description}

In this appendix, we present algorithmic details of the three sets of kernels developed for {\tt e3j}: CUDA, Pallas GPU, and Pallas TPU. The CUDA kernel offers the broadest applicability across various GPU architectures and is fully deterministic, while Pallas GPU focuses on performance for the latest GPU architectures and JAX versions. We first present the relevant notation and recall the operation that needs to be performed as part of the Clebsch--Gordan Tensor Product and associated message passing, and then present in order the CUDA, Pallas GPU and Pallas TPU kernels. In order to facilitate understanding for readers with different backgrounds, we have added a short section outlining the key hardware concepts for GPU and TPU in appendix \ref{apx:hardware_concepts}.

\subsection{Mathematical Details} \label{sec:maths}

In this section we detail the notation used and provide a mathematical description for the operation performed by the kernel. For further details on the motivation for the construction of these operations from a theoretical standpoint, please refer to the original literature. 

\paragraph{Notation.}

\begin{itemize}
    \item ${\bf x}$: Left-hand-side features. In the Tensor Product, it is an arbitrary feature tensor, the Message Passing operation, it represents the sender node features. It is indexed along three axes: batch elements (usually nodes), equivariant features (see below for indexing notations), and channels
    \item ${\bf y}$: Right-hand-side features. In the Tensor Product, it is an arbitrary feature tensor, in the Message Passing operation, it usually represents (broadcasted) spherical harmonics embeddings. It is indexed along three axes: batch elements (usually edges), equivariant features (see below for indexing notations), and channels.
    \item ${\bf s}$: Edge scalars. Edge specific weighting, usually computed using a MLP on radial embedding projecting to the number of channels, for each edge. It is indexed along edges, equivariant features, and channels.
    \item ${\bf m}_b$: Aggregated messages on receiver node. The output of the message passing convolution for each node. It is worth noting that it does not int principle,  have the same feature dimension as ${\bf x}$ as multiplicities arise during the tensor product (these are in general contracted back to the number of channels in channel mixing). It is likewise indexed by nodes, equivariant features, and channels. 
    \item $C$: Clebsch--Gordan coefficients. Its values are indexed by outputs ($i0$), l.h.s input ($i1$) and r.h.s. input ($i2$) feature indexing (each $i0$, $i1$, $i2$ represents a $\ell$, $m$ combination) .
    \item ${\bf m}_{ab}$: Edge level message between a sender node and a receiver node.
    \item $a$: Sender indexing
    \item $b$: Receiver indexing
    \item $q$: Channel indexing
    \item $p$: sparse Clebsch--Gordan record index (when unrolled)
    \item $N_q$: The number of channels
    \item $N_p$: The number of non-zero Clebsch--Gordan coefficients
    \item $c_p = C^{i0_p}_{i1_p,i2_p}$: Coefficient value of record $p$ 
\end{itemize}

\paragraph{Mathematical operations.}

Following the notation above, we can outline the operations performed by the kernels:

\begin{itemize}
    \item \textbf{Tensor product of geometric features:} This formula generalises to bilinear operation on arbitrary feature vectors, it includes the Clebsch--Gordan tensor products when appropriate weights are selected. Here we treat the sender / receiver indices as implicit as this operation can be performed arbitrarily on any batch element (i.e. nodes or edge features). ${\bf z}$ is an arbitrary notation to represents the output feature tensor following a tensor product. 
    \begin{equation} \label{eq:tensor_product_formula}
        {\bf z}^{i0,q} = \sum_{i1, i2} 
        C^{i0}_{i1,i2} * {\bf x}^{i1, q} * 
        {\bf y}^{i2, q}
    \end{equation}
    \item \textbf{Message weighting with edge scalars:} From the computed Tensor Product features on a given edge, one can weight the edge specific message before aggregation. This is usually done through a MLP mapping radial embeddings to a set number of channels. In the notation below, the index $i3$ maps the feature coordinate $i0$ to its irreducible block (piecewise-constant on irreducible subspaces), 
    we write this dependency as $i3(i0)$ for simplicity. The mapping from feature coordinates to scalar indices is performed at coefficient construction time from I/O representations. This operation can be viewed as a Tensor Product with the r.h.s. input being scalars instead of arbitrary geometric features. 
    \begin{equation} \label{eq:edge_scalar_weighting}
        {\bf m}^{i0,q}_{ab} = {\bf z}^{i0,q}_{ab}
         * {\bf s}^{i3(i0), q}_{ab}
    \end{equation}
    \item \textbf{Message Passing Convolution:} 
    The complete operation for the message passing convolution, per node, can be written concisely 
    as this common particular version of \eqref{eq:message-passing-linear}:
\begin{equation}\label{eq:message-passing}
    {\bf m}_{b} = \sum_{a \sim b} {\bf \tilde m}_{ab}
    \quad{\rm where}\quad
    {\bf \tilde m}_{ab} = {\bf s}_{ab} \cdot ({\bf x}_a \otimes {\bf y}_{ab})
\end{equation}
given node features ${\bf x}_a$, edge features ${\bf y}_{ab}$, scalar embeddings ${\bf s}_{ab}$ 
and letting the dot denote the scalar mixing operation. Expanding equations \eqref{eq:tensor_product_formula} and \eqref{eq:edge_scalar_weighting} yields
    \begin{equation} \label{eq:message_passing_convolution}
        {\bf m}_b^{\tt i0,q} = \sum_{a \sim b} 
        \sum_{i1, i2}
        C^{i0}_{i1,i2} \, {\bf x}_a^{\tt i1, q} \,
        {\bf y}_{ab}^{\tt i2, q} \, {\bf s}_{ab}^{\tt i3(i0), q}.
    \end{equation}
The equivalence between \eqref{eq:message-passing} and \eqref{eq:message_passing_convolution}, 
reflecting the associativity of bilinear couplings, leads to different computation graphs and implementations, as depicted in figure \ref{fig:convolution-graph}.
\end{itemize}

\paragraph{Differentiation.}
All of our kernels and associated primitives are infinitely differentiable. We outline below the mathematical derivation of their so-called reverse-mode AD primitives, or vector-jacobian products (VJP) rules, obtained by first differentiating the smooth function above 
a set of inputs, called {\em primals}, before transposing the linearized map. 
The transposed differential acts on {\em cotangents} (linear forms on tangent vectors) 
in the reversed direction, i.e. it maps {\em output cotangents} to {\em primal cotangents} so as 
to enable back-propagation of gradients in a complete neural network architecture.

\begin{figure}[ht]
    \centering
    \includegraphics[width=1\linewidth]{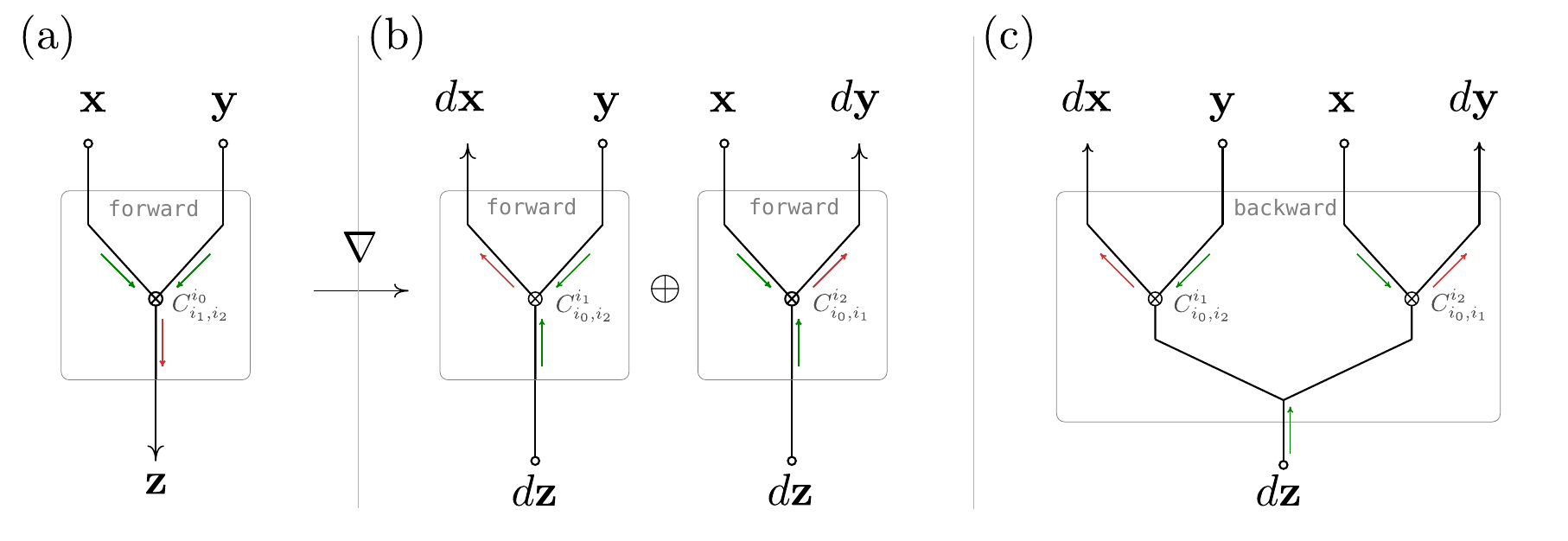}
    \caption{\textbf{Computation graphs for the tensor product forward pass (a) and backward pass (b-c).}
    Back-propagating through a tensor product operation yields two transposed tensor product operations, 
    by the Leibniz rule. Instead of calling two tensor product kernels, a dedicated backward kernel 
    reuses device code for batch-wise bilinear coupling, while streaming through batches of the $d{\bf z}$ 
    cotangents only once and thus saving memory traffic.
    }
    \label{fig:back_cuda_tp}
\end{figure}

\begin{itemize} 
\item \textbf{Tensor product backward:} 
By bilinearity of the tensor product, the Leibniz rule gives the forward-mode AD rule 
for ${\bf z} = {\bf x} \otimes {\bf y} \in X\otimes Y$ as: 

\begin{equation} \label{eq:tp-fwd}
    \delta {\bf z} = \delta {\bf x} \otimes {\bf y} + {\bf x} \otimes \delta{\bf y}
\end{equation}

where $\delta {\bf x} , \delta {\bf y}$ denote {\em tangent vectors} of $T_{\bf x}  X$ and $T_{{\bf y}} Y$ respectively, mapped to an output tangent $\delta {\bf z} \in T_{\bf z}(X \otimes Y)$ by the linearized 
tensor product operation. Equation \eqref{eq:tp-fwd} thus defines a {\em linear map} 
$L_{{\bf x},{\bf y}} : T_{\bf x} X \oplus T_{\bf y} Y \to T_{\bf z}(X \otimes Y)$ such that 
$\delta {\bf z} = L_{\bf x, y}(\delta {\bf x}, \delta {\bf y})$. 

The backward tensor product primitive is the adjoint (a.k.a. transpose)\footnote{
    Formally, the adjoint $L_{\bf x, y}^*$ is the composition or {\em pull-back} 
    $d {\bf z} \circ L_{{\bf x} ,{\bf y}}$. Composing the linear form $d {\bf z}$ with 
    the linear map $L_{{\bf x} ,{\bf y}}$ defines a linear form on $T_{\bf x}X\oplus T_{\bf y}Y$, 
    or two linear forms on $T_{\bf x}X$ and $T_{\bf y} Y$, a.k.a. {\em cotangents}.
} 
of $L_{{\bf x}, {\bf y}}$, which maps an output cotangent $d {\bf z}\in T^*_{\bf z}(X \otimes Y)$ to primal
cotangents $d {\bf x} \in T^*_{\bf x} X$ and  $d{\bf y} \in T^*_{\bf y}Y$. In practice, this means transposing the partially applied tensor product maps ${\bf x} \otimes -$ and 
$- \otimes {\bf y}$, respectively consuming $\delta {\bf y}$ and $\delta {\bf x} $ in \eqref{eq:tp-fwd}. 
The transposition amounts to applying a permutation of indices. For $d\bf x$, each record is permuted as $(i0,i1,i2)\mapsto(i1,i2,i0)$ and sorted by the new output index $i1$; for $d\bf y$ we use $(i0,i1,i2)\mapsto(i2,i1,i0)$ and sort by the new output index $i2$. The coefficient values are reordered identically. 

Writing ${\bf z} = {\tt tensor\_product(C^{i0}_{i1 i2}, {\bf x}, {\bf y})}$ in reference to algorithm \ref{alg:tp-trailing}, the source cotangents can be evaluated with the same sparse kernel after permuting the coefficient records: 
\begin{equation} \label{eq:tp-bwd}
    \left\{
    \begin{array}{ccc}
        d {\bf x} &= &{\tt tensor\_product(C^{i1}_{i2 i0}, \bf y}, d{\bf z}) \\
        d {\bf y} &= &{\tt tensor\_product(C^{i2}_{i1 i0}, \bf x}, d{\bf z})
    \end{array}
    \right.
\end{equation}
The operation is also illustrated in Fig. \ref{fig:back_cuda_tp}.

Although this simple backward algorithm provides infinite differentiability, 
it has the disadvantage of loading output cotangents $\delta {\bf z}$ twice in on-device memory, which can be significantly expensive as the output feature dimensions scales as the product of input feature dimensions.

\item{\textbf{Message-passing backward:}} By duality, gather and scatter-add operations are swapped in reverse-mode differentiation and the backward message-passing rule can be expressed as a message-passing operation on the transposed graph, with multiple bilinear couplings being performed. 
Letting ${\bf z}_{ab} = {\bf x}_a \otimes {\bf y}_{ab}$ in \eqref{eq:message-passing}, input cotangents may be computed from receiver cotangents $d{\bf m}_b$ as: 
\begin{equation}
    {\tt conv\_bwd({\bf x, y, s, d{\bf m}})} =
    \left\{\begin{array}{ll} \label{eq:conv-bwd}
        d{\bf x}_a &= 
        \displaystyle
        \sum_{b \sim a} {\bf s}_{ab} \cdot  (
        {\bf y}_{ab} \otimes d{\bf m}_b) \\[.7em]
        d{\bf y}_{ab} &= 
        \displaystyle
        {\bf s}_{ab} \cdot ({\bf x}_{a} \otimes d{\bf m}_b ) \\[.7em]
        d{\bf s}_{ab}^{(l)} &= \displaystyle
        \sum_{m=-l}^{+l} {\bf z}_{ab}^{(l,m)} 
        \cdot d{\bf m}_b^{(l,m)}
    \end{array}\right.
\end{equation}
Because each cotangent is computed by an ad-hoc operation, \eqref{eq:conv-bwd} may hardly reuse 
device code from the forward pass yet specialized backward device code can prove more efficient\footnote{
    Our CUDA kernels prioritize code reuse while our Pallas kernels use specialized code 
    in backward operations, see below.
}. 
Instead of \eqref{eq:conv-bwd}, if one views the message-passing as a trilinear coupling 
-- see figure \ref{fig:convolution-graph} and equation \eqref{eq:message-passing} -- 
the VJP can be expressed as three distinct trilinear coupling calls to 
a single \texttt{bigotimes} routine, via a straightforward 
generalization of \eqref{eq:tp-bwd} to three operands.

\item \textbf{Message-passing double backward:} 
Deriving the second order rule from \eqref{eq:conv-bwd} is best viewed through the diagram \ref{fig:convolution-graph} 
back-propagated twice, each differentiation of \texttt{bigotimes} 
yielding three transposed \texttt{bigotimes} calls writing to each of the input leaves.
Denoting by $\delta d{\bf x}, \delta d{\bf y}, \delta d {\bf s}$ the second-order variations 
of the primal inputs (first-order variations of cotangents returned by the backward pass), 
the associated variations $\delta {\bf x}, \delta{\bf y}, \delta {\bf s}, \delta d{\bf m}$ 
returned by the backward rule are defined in terms of lower-order primitives as: 

\begin{equation}\label{eq:conv-bwd-bwd-ddm}
\delta d {\bf m} = 
{\tt conv}(\delta d{\bf x}, {\bf y}, {\bf s}) + 
{\tt conv}({\bf x}, \delta d{\bf y}, {\bf s}) + 
{\tt conv}({\bf x}, {\bf y}, \delta d{\bf s})
\end{equation}

\begin{equation} \label{eq:conv-bwd2-xys}
\left\{ \begin{array}{l}
\delta {\bf x} = \delta_{y}{\bf x} + \delta_{s}{\bf x} \\
\delta {\bf y} = \delta_{x}{\bf y} + \delta_{s}{\bf y} \\
\delta {\bf s} = \delta_{x}{\bf s} + \delta_{y}{\bf s} 
\end{array}\right.
\quad
{\rm where}\quad \left\{ \begin{array}{lll}
{\rm - }, \delta_x {\bf y}, \delta_x{\bf s} &= 
{\tt conv\_bwd}(\delta d{\bf x}, {\bf y}, {\bf s}, d{\bf m})  \\ 
\delta_y {\bf x},{\rm -}, \delta_y{\bf s} &= 
{\tt conv\_bwd}({\bf x}, \delta d {\bf y}, {\bf s}, d{\bf m}) \\
\delta_s {\bf x}, \delta_s{\bf y},{\rm -} &= 
{\tt conv\_bwd}({\bf x}, {\bf y}, \delta d{\bf s}, d{\bf m}) \\
\end{array}\right. .
\end{equation}
\end{itemize}
Note that defining AD rules via cyclic references to lower-order differentials, 
as \eqref{eq:tp-bwd} and (\ref{eq:conv-bwd-bwd-ddm}-\ref{eq:conv-bwd2-xys}) do,
automatically yields infinitely differentiable JAX primitives. 
At this time {\tt e3j} doesn't define double-backward kernels, in contrast with 
\texttt{OpenEquivariance} \citep{openeq}.
Double-backward kernels could save unused cotangents from being computed in \eqref{eq:conv-bwd2-xys},
or save memory traffic by streaming through the primals in a single loop.
Due to the increased number of I/O arrays, a GPU implementation would 
likely rely on implicit L1/L2 caching over shared memory buffering, 
yet these constraints do not apply on TPU whose VMEM slice is much larger. 
Double-backward optimization is left for future work.

\subsection{CUDA kernels}

A significant design choice of {\tt e3j} is to rely on a sparse and agnostic representation of total Clebsch-Gordan coefficients, in which equivariant coordinates (traditionally a set of triplets $(\kappa, \ell, m)$ for the output and each of the two inputs, hence a total of 9 integers) are mapped to unfolded feature coordinates 
$i0, i1, i2$, for output, l.h.s input and r.h.s input, running over the whole I/O dimensions. 
Melding the equivariant coordinates together into opaque indices means we can rely on a very generic sparse tensor product algorithm, suitable for different applications.

The number of irreducible blocks in the sum scales as $O(\lmax^3)$ -- number of input degree pairs multiplied by the number of possibilities for the output degree --  
while the number of non-zero coefficients within each block scales as $\lmax^2$, due to the spin constraints $M = m + m'$. Since the feature dimension $D$ scales as $\lmax^2$, and our algorithm \ref{alg:tp-trailing} scales as $O(\lmax^5) = O(D^{5/2})$ in terms of floating point operations (FLOPs). In practical situations we targeted, $\lmax$ is a fixed hyperparameter of the MLIP model and the number of non-zero coefficients is a few hundreds or less. 

\subsubsection{CUDA Tensor Product operation}

The main bottleneck of the sparse, JAX-based implementation of the Tensor Product
is the final scatter-reduction step, which scaled poorly to large input sizes. 
Our CUDA algorithm bypasses this bottleneck and implements a number of focused optimization, the key differentiating implementation factors of algorithm \ref{alg:tp-trailing} are explained below: 

\begin{itemize}
    \item \textbf{Trailing channels:} we avoid any inter-thread communication by sequentially reducing the feature axis, and parallelizing across the trailing channel dimension instead. The 32 threads of a warp (simultaneously scheduled to execute the same instruction) can thus process distinct channels in parallel, while doing the same work: i.e. warp lanes process distinct channels while traversing the same ordered sequence of CG coefficient records.
    \item \textbf{Vectorization:} processing 2 or 4 channels simultaneously inside the inner coefficient loop, 
    so as to amortize by the same factor the cost of coefficient loads.
    Ablation studies reveal using wider \texttt{float2} to \texttt{float4} datatypes in the coefficient loop 
    nearly double the throughput compared to the non-vectorized kernel, as can be expected from the shared memory 
    traffic per operand coupling: $(2 \times 4 + 16) = 24$ bytes loaded per batch, 
    versus $(4 \times 2 \times 4 + 16)/4  = 12$ bytes loaded per batch with 4-fold vectorization, with 16B coefficients.
    \item \textbf{Coefficient packing:}  using narrow index data types so as to halve the size of each coefficient load when dimensions are small enough. The array of coefficients ${\tt C}$ is passed as a pointer to ${\tt Coef}$ structures holding one 32-bit {\tt float} value and three indices of type ${\tt Idx}$, where: 
        \begin{itemize}     
            \item ${\tt Idx = uint8}$ if all I/O dimensions are bounded by 255. ${\tt Coef}$ is then a 56 bit type, aligned to 64 bits, which can be loaded in a single ${\tt LDS.64}$ instruction. 
            \item ${\tt Idx = int32}$ by default. ${\tt Coef}$ is then a 128 bit type, which can 
            be loaded in a single ${\tt LDS.128}$ instruction. 
    \end{itemize}
    In practice, narrow indices have to be expanded to 32 bits in CUDA registers, and 
    the upside of index narrowing is mostly that of a reducing shared memory pressure.
    \item \textbf{Coefficient distribution and occupancy:} our kernels buffer input rows in shared memory to 
    avoid long-scoreboard stalls when loading operands during the inner loop of algorithm \ref{alg:tp-trailing}. 
    Since SMEM footprint limits the number of resident blocks on each SM, maximizing the number of resident threads 
    requires increasing the block sizes above the given channel count. To that end, the coefficient loop is parallelized 
    along the y-axis of a block, by first grouping coefficients by output indices before splitting them in evenly-sized groups.
\end{itemize}

\begin{algorithm} 
    \caption{
        Tensor product evaluation: parallel trailing channels, sequential aggregation
    }\label{alg:tp-trailing}
    \KwData{non-zero COO coefficients 
    ${\tt C} = ({\tt i0, i1, i2, val}) \in (\N^3 \times \R)^{N_c}$ sorted by output index,
        l.h.s. ${\bf x} \in \R^{d \times N_q}$, 
        r.h.s. ${\bf y} \in \C^{d' \times N_q}$, 
        thread index $t \in \N$\\
    }
    \KwResult{output ${\bf z} = {\bf x} \otimes {\bf y} \in \R^{D}$} 
    ${\tt i0} \gets 0$ \;
    ${\tt z_{i0}} \gets 0$\;
    \For{$p = 0 \dots N_c - 1$}{
        ${\tt coef} \gets {\tt LOAD} \: {\tt C}[p]$ \;
        ${\tt i1, i2} \gets {\tt coef.i1}, {\tt coef.i2}$ \;
        ${\tt x_{i1}} \gets {\tt LOAD} \: {\bf x}{\tt [i1, t]}$ \;
        ${\tt y_{i2}} \gets {\tt LOAD} \: {\bf y}{\tt[i2, t]}$ \;
        \If{${\tt coef.i0} == {\tt i0}$} {
            ${\tt z_{i0}} += {\tt coef.val * x_{i1} * y_{i2}}$ \; 
        } \Else { 
            ${\bf z}{\tt[i0,t]} \gets {\tt STORE} \: {\tt z_{i0}}$ \;
            ${\tt z_{i0}} \gets {\tt coef.val * x_{i1} * y_{i2}}$ \; 
            ${\tt i0} \gets {\tt coef.i0} $
        }
    }       
\end{algorithm}

\begin{comment}
\begin{algorithm}[ht]
    \caption{
        Tensor product evaluation: parallel trailing channels, sequential aggregation
    }
    \label{alg:tp-trailing}
    \KwData{
        non-zero coefficient records
        ${\tt C}[p]=(i0_p,i1_p,i2_p,c_p)$, sorted by $i0_p$,
        where $c_p=C^{i0_p}_{i1_p,i2_p}$;
        l.h.s.\ ${\bf x} \in \C^{\dots \times d \times N_q}$;
        r.h.s.\ ${\bf y} \in \C^{\dots \times d' \times N_q}$;
        channel $q$
    }
    \KwResult{
        output ${\bf z} = {\bf x} \otimes {\bf y}
        \in \C^{\dots \times D \times N_q}$
    }

    ${\tt i0} \gets i0_0$\;
    ${\tt z} \gets 0$\;  

    \For{$p = 0,\ldots,N_{ p}-1$}{
        ${\tt coef} \gets {\tt LOAD}\ {\tt C}[p]$\;
        ${\tt x} \gets {\tt LOAD}\ {\bf x}^{i1_p,q}$\;
        ${\tt y} \gets {\tt LOAD}\ {\bf y}^{i2_p,q}$\;

        \If{$i0_p = {\tt i0}$}{
            ${\tt z}
            \mathrel{+}=
            c_p\,{\tt x}\,{\tt y}$\;
        }
        \Else{
            ${\bf z}^{{\tt i0},q}
            \gets {\tt STORE}\ {\tt z}$\;

            ${\tt z}
            \gets
            c_p\,{\tt x}\,{\tt y}$\;

            ${\tt i0} \gets i0_p$\;
        }
    }

    ${\bf z}^{{\tt i0},q}
    \gets {\tt STORE}\ {\tt z}$\;
\end{algorithm}
\end{comment}

\paragraph{Backward pass.} Back-propagating through the ${\tt tensor\_product}$ primitive can be implemented as two additional calls to the same ${\tt tensor\_product}$ kernel: hence yielding $3n$ kernel calls for a model with $n$ tensor products. This is made possible by the fact that algorithm \ref{alg:tp-trailing} is generic with respect to the sparse coefficient array, see equation \eqref{eq:tp-bwd}.

Although this simple backward algorithm provides infinite differentiability, 
it has the disadvantage of loading output cotangents $\delta {\bf z}$ twice in on-device memory, which can be significantly expensive as the output feature dimensions scales as the product of input feature dimensions.
Our backward tensor product kernel streams through batches of $({\bf x}, {\bf y}, d{\bf z})$ 
once and computes the two cotangents as \eqref{eq:tp-bwd}, by reusing the same device code for the bilinear 
coupling (algorithm \ref{alg:tp-trailing}).

\subsubsection{CUDA Message Passing Convolution} \label{apx:convolution}

It is worth noting that our CUDA message passing kernel is not constrained to edge based spherical harmonic embedding as second operand. Many Euclid equivariant MLIP architectures use this message-passing update of node features, such as MACE \citep{MACE} and NequIP~\citep{Nequip}, and
message formation is a known bottleneck of most MLIP architectures.\footnote{
    MACE can be a notable exception due to its expensive multi-body atomic cluster expansion. 
    However, with correlation 3 and maximal degree 1 for output node features, 
    about 80\% of runtime can be spent on the message-passing operation.
} For practical improvements, it may however be more important to consider hardware behavior and constraints at typical values of $\lmax$ rather 
than complexity arguments on hyperparameters. 

Even with an ideal memory-bound tensor product kernel, the message-passing operation may imply materializing edge features $\tilde {\bf m}_{ab} = {\bf x}_a \otimes {\bf Y}_l({\bf r}_{ab})$ in global memory. Their leading axis, the number of edges, is more than one order of magnitude larger than the number 
of atoms $N$ (average number of neighbors of around 45 in organic molecules at {5\AA} cutoff). 

\begin{figure}[ht]
    \centering
    \includegraphics[width=0.8\linewidth]{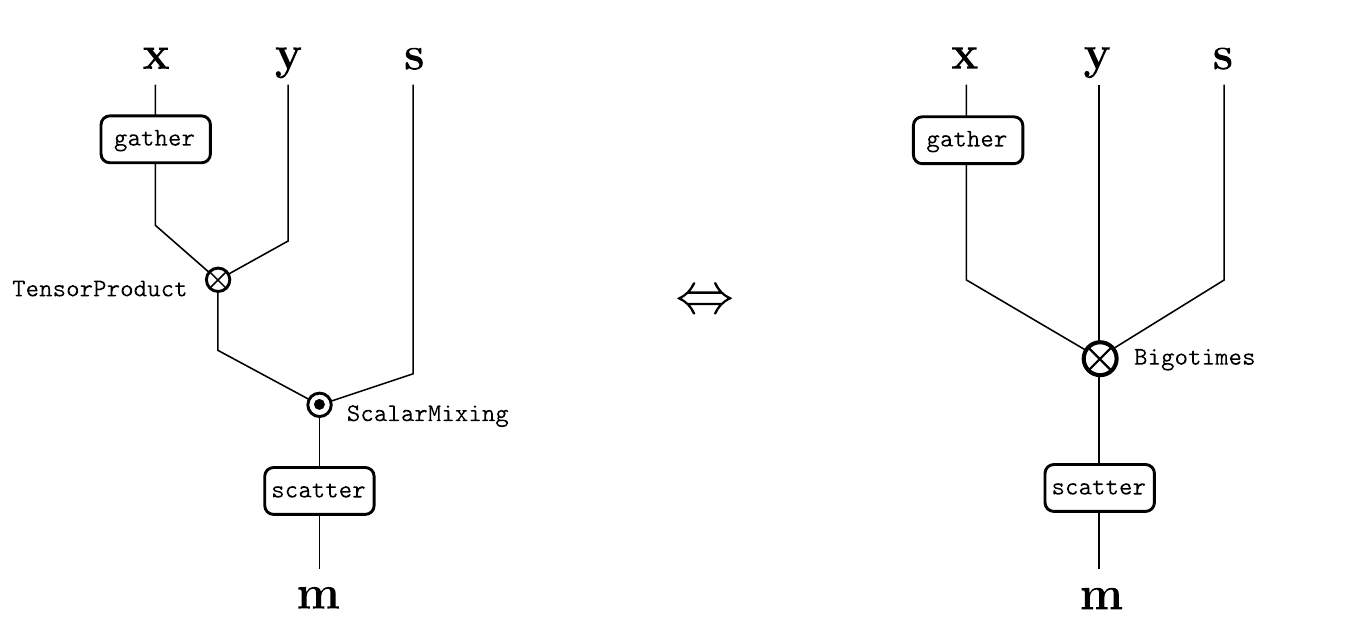}
    \caption{
    \textbf{Equivalent computation graphs for the convolution operation:} the edge-wise composition of a scalar mixing on top of a bilinear      tensor product (left) is equivalent to an edge-wise trilinear coupling (right) by associativity. 
    }
    \label{fig:convolution-graph}
\end{figure}

Our CUDA convolution kernel streams through edges and computes messages via a straightforward 
overload of algorithm \ref{alg:tp-trailing} to three operands given sparse 4D coefficients 
$C^{i0}_{i1i2i3}$, 
which we refer to as the \texttt{bigotimes} device routine. See equation \eqref{eq:message_passing_convolution} and figure \ref{fig:convolution-graph}.

The trilinear mixing is embedded within an outer loop over receiver nodes, and an inner loop over 
sender nodes (neighbors) which relies on a Compressed Sparse Row (CSR) representation of the 
adjacency matrix. During each neighbor loop, messages $\tilde {\bf m}_{ab}$ 
are accumulated in a shared memory buffer. 
While this accumulation strategy and CSR adjacency format 
enables efficient and deterministic message aggregation (free of compare-and-swap operations, 
a.k.a. {\em atomics}), the additional operand and message buffers
increase the shared memory pressure compared to the raw tensor product kernel.

\paragraph{Backward pass.} The current backward kernel processes each of the trilinear mixing operations of Eq. \eqref{eq:conv-bwd} with a common {\texttt{bigotimes()}} routine (see Fig. \ref{fig:convolution-graph}). While postponing the scalar mixing step could save a few FMUL/FMA operations in the forward pass, treating scalar mixing as a distinct operation in the backward pass leads to more complex device code and accumulation patterns (inner product reduction on channels of $d{\bf y}_{ab}$) and increased register pressure. While backward convolution still shows a good margin for optimization, relative algorithmic simplicity remains a constraint for high enough occupancy on the device.

\subsection{Pallas GPU convolution kernel}
\label{sec:mosaic-gpu-convolution}

The Pallas GPU kernel implements the same message-passing operation as Eq.~\ref{eq:message-passing}, but differs from the CUDA implementation in how the sparse Clebsch--Gordan contraction and edge traversal are scheduled. Its main distinguishing feature is trace-time specialization of the Clebsch--Gordan coefficients. Edges are additionally traversed in receiver order, which trades receiver-local, atomics-free forward aggregation against locality and cache reuse in computations where atomic accumulation is required. Algorithm~\ref{alg:mosaic-gpu-conv} summarizes the forward kernel. Note that buffering of ${\bf x}_a$, ${\bf y}_{ab}$ and ${\bf s}_{ab}$ through 
SMEM is kept implicit for conciseness, and the {\tt LOAD} directive refers to a shared memory load ({\tt LDS}) as in algorithm \ref{alg:tp-trailing}.

\paragraph{Trace-time Clebsch--Gordan specialization.} Unlike the AOT-compiled CUDA Tensor Product implementation of Algorithm~\ref{alg:tp-trailing}, the Pallas kernel does not loop through an array of coefficients at runtime. 
The non-zero coefficients and their indices depend only on the selected
I/O feature spaces, and are therefore statically available for the Mosaic compiler\footnote{
    Clebsch-Gordan coefficients grouped by indices are actually passed as nested Python dictionaries in the kernel source, which forces inlining by the upstream HLO compiler.
}. Writing the sparse records as \[ (i0_p,i1_p,i2_p,c_p), \] the loop over $p$ is unrolled at trace time and the corresponding contraction paths are embedded directly in the generated program. There is therefore no 
{\tt LOAD} of coefficients and their indices at runtime, in contrast 
with Algorithm \ref{alg:tp-trailing}, and the operand values can be loaded 
at each iteration without waiting for the coefficient and its indices to become 
available in registers.

The records are grouped by output index $i0$, and passed as a Python dictionary -- which forces inlining by the upstream HLO compiler -- mapping $i0$ to 
a sequence ${\cal C}[{\tt i0}]$ of triplets $(i1_p, i2_p, c_p)$ in Algorithm~\ref{alg:mosaic-gpu-conv}. For a fixed edge, output feature $i0$, and channel $q$, the contraction \[ \sum_{i1,i2} C^{i0}_{i1,i2}\, {\bf x}^{i1,q}_a\, {\bf y}^{i2}_{ab} \] can consequently be accumulated in registers before applying the corresponding edge scalar ${\bf s}^{i3(i0),q}_{ab}$. Trace-time specialization removes the coefficient loads and run-time loop and indexing overhead associated with traversing the sparse representation. This leads to having mostly {\tt FMA} and {\tt LDS} instructions that are all visible to the compiler, and can be reordered opportunistically to hide 
the latency of these instructions.

\begin{algorithm}[ht]
\caption{
Pallas GPU message-passing convolution: receiver-ordered aggregation and trace-time Clebsch--Gordan specialization.
}
\label{alg:mosaic-gpu-conv}
\DontPrintSemicolon
\KwData{static coefficients grouped by output index
${\cal C}[{\tt i0}] = ({\tt i1}_p, {\tt i2}_p, c_p)_{p \in {\cal P}_{\tt i0}}$;
l.h.s.\ node features
${\bf x}\in\R^{N\times d\times N_q}$;
r.h.s.\ edge features
${\bf y}\in\R^{E\times d'}$;
edge scalars ${\bf s} \in \R^{E \times d'' \times N_q}$;
sender indices ${\tt sender}$;
receiver CSR pointers ${\tt rowptr}$;
channel $q$
}
\KwResult{
aggregated messages
${\bf m}\in\R^{N\times D\times N_c}$
}

\ForEach{receiver $b$}{
    $e_{\rm begin}, e_{\rm end} \gets{\tt rowptr}[b], {\tt rowptr}[b+1]$\;

    ${\tt acc}\gets0$\;
    %\tcp{receiver accumulator remains in registers}

    \For{$e=e_{\rm begin},\ldots,e_{\rm end}-1$}{
        $a\gets {\tt sender}[e]$\;        
        \tcp{UNROLL}
        \For{${\tt i0} = 0,\ldots,D - 1$}{
            ${\tt z}\gets0$\;
            \For{$({\tt i1}_p, {\tt i2}_p, c_p) \in {\cal C}[{\tt i0}]$}{
            ${\tt x}\gets{\tt LOAD}\ {\bf x}_a[{\tt i1}_p,q]$\;
            ${\tt y}\gets{\tt LOAD}\ {\bf y}_{ab}[{\tt i2}_p]$\;
            ${\tt z}
            \mathrel{+}=
            c_p * {\tt x} * {\tt y}$\;
            }
            
                ${\tt s}\gets{\tt LOAD}\ {\bf s}_{ab}[i3({\tt i0}),q]$\;

                ${\tt acc}[{\tt i0}, q]
                \mathrel{+}=
                {\tt s} * 
                {\tt z}$\;
            
        }
    }

    ${\bf m}_b
    \gets{\tt STORE}\ {\tt acc}$\;
}
\Return{${\bf m}$}
\end{algorithm}

\paragraph{Receiver-ordered aggregation.} Edges are stored in compressed sparse row (CSR) format ordered by receiver $b$, such that $[{\tt rowptr}[b],{\tt rowptr}[b+1])$ contains all edges incident on $b$. In the forward pass, one cooperative thread array (CTA) processes the complete reduction for a receiver. Its accumulator is initialized in registers, updated while traversing the incoming edges, and written to GMEM after the final edge. 

This avoids inter-CTA atomic aggregation of ${\bf m}_b$. In an edge-parallel implementation, several CTAs may contribute simultaneously to \[ {\bf m}_b \mathrel{+}= {\bf \tilde m}_{ab}, \] requiring atomic read--modify--write operations. Receiver-stationary aggregation instead keeps the partial sum private to one CTA, providing a fixed receiver-local accumulation order and avoiding repeated stores of partial results. 

The same property does not hold generally in the backward kernel. Receiver ordering makes consecutive edges reuse the receiver cotangent $d{\bf m}_b$, improving temporal locality and potentially L2-cache reuse, but gradient contributions need not remain local to one receiver. The implementation therefore uses atomic accumulation where multiple CTAs contribute to the same $d{\bf x}$, $d{\bf y}$, or $d{\bf s}$ output. Receiver ordering thus reflects a trade-off between atomics-free, deterministic receiver aggregation in the forward pass and improved locality in computations whose output aggregation may require atomics.

\paragraph{Edge traversal.}
The receiver CSR interval is traversed sequentially, or in small contiguous tiles. Pallas's \texttt{emit\_pipeline} can stage receiver-ordered edge-local operands from GMEM to SMEM using asynchronous transfers, including TMA, while the current tile is processed. The sender access ${\bf x}_a$ remains an indirect gather.

This pipelining is a scheduling optimization rather than a defining feature of the kernel: its benefit depends on the balance between memory traffic and tensor-product computation. The central consequences of receiver ordering are instead the forward aggregation strategy and the locality obtained when repeatedly accessing receiver-associated quantities.

\begin{algorithm}[ht]
\caption{
Pallas GPU message-passing convolution, backward pass: receiver-ordered traversal and trace-time Clebsch-Gordan specialization.
}
\label{alg:mosaic-gpu-conv-bwd}
\DontPrintSemicolon
\KwData{static coefficients grouped by output index
${\cal C}[{\tt i0}] = ({\tt i1}_p, {\tt i2}_p, c_p)_{p \in {\cal P}_{\tt i0}}$;
l.h.s.\ node features
${\bf x}\in\R^{N\times d\times N_q}$;
r.h.s.\ edge features
${\bf y}\in\R^{E\times d'}$;
edge scalars ${\bf s} \in \R^{E \times d'' \times N_q}$;
output cotangent $d{\bf m}\in\R^{N\times D\times N_q}$;
sender indices ${\tt sender}$;
receiver CSR pointers ${\tt rowptr}$;
channel $q$
}
\KwResult{
input cotangents
$d{\bf x}\in\R^{N\times d\times N_q}$,
$d{\bf y}\in\R^{E\times d'}$,
$d{\bf s}\in\R^{E\times d''\times N_q}$
}

$d{\bf x}\gets0$\;

\ForEach{receiver $b$}{
    $e_{\rm begin}, e_{\rm end} \gets{\tt rowptr}[b], {\tt rowptr}[b+1]$\;

    \For{$e=e_{\rm begin},\ldots,e_{\rm end}-1$}{
        $a\gets {\tt sender}[e]$\;
        ${\tt dx}, {\tt dy}, {\tt ds}\gets0$\;

        \tcp{UNROLL}
        \For{${\tt i0} = 0,\ldots,D - 1$}{
            ${\tt s}\gets{\tt LOAD}\ {\bf s}_{ab}[i3({\tt i0}),q]$\;
            ${\tt dm}\gets{\tt LOAD}\ d{\bf m}_b[{\tt i0},q]$\;
            ${\tt dz}\gets{\tt s} * {\tt dm}$,
            \quad
            ${\tt z}\gets0$\;
            \For{$({\tt i1}_p, {\tt i2}_p, c_p) \in {\cal C}[{\tt i0}]$}{
                ${\tt x}\gets{\tt LOAD}\ {\bf x}_a[{\tt i1}_p,q]$\;
                ${\tt y}\gets{\tt LOAD}\ {\bf y}_{ab}[{\tt i2}_p]$\;
                ${\tt z} \mathrel{+}= c_p * {\tt x} * {\tt y}$\;
                ${\tt dx}[{\tt i1}_p] \mathrel{+}= c_p * {\tt y} * {\tt dz}$\;
                ${\tt dy}[{\tt i2}_p] \mathrel{+}= c_p * {\tt x} * {\tt dz}$\;
            }
            ${\tt ds}[i3({\tt i0})] \mathrel{+}= {\tt dm} * {\tt z}$\;
        }

        $d{\bf s}_{ab}[:,q] \gets{\tt STORE}\ {\tt ds}$\;
        $d{\bf y}_{ab} \gets{\tt STORE}\ \textstyle\sum_q {\tt dy}$
        \tcp*{reduction over channels}
        $d{\bf x}_a \gets{\tt ATOMIC\_ADD}\ {\tt dx}$\;
    }
}
\Return{$d{\bf x}, d{\bf y}, d{\bf s}$}
\end{algorithm}

\subsection{Pallas TPU kernel}
\label{apx:convolution:tpu}

\paragraph{Notation}

The following notation is specific to the TPU implementation: 
\begin{itemize}
    \item $B$: number of edges in an edge tile 
    \item $t$: edge-tile index 
    \item ${\bf a}_t,{\bf b}_t$: sender and receiver index vectors for tile $t$ 
    \item ${\bf y}_t,{\bf s}_t$: edge-feature and edge-scalar tiles 
    \item $\hat{\bf x}_t={\bf x}[{\bf a}_t]$: sender features gathered for tile $t$
    \item $k=i3(i0)$: scalar-mixing block of output coordinate $i0$, which selects the edge scalar ${\bf s}[k]$
    \item $\mathit{acc}$, $\mathit{current}$ and $\mathit{stage}$: state used by the segmented receiver or sender reduction. 
\end{itemize} 

The following acronyms can also be useful, though for more details we refer readers to Appendix \ref{apx:hardware_concepts}: HBM denotes the TPU's high-bandwidth memory, VMEM its vector memory, and SMEM its scalar memory.

\paragraph{Tiling over edges rather than receivers.}
In contrast to the receiver-stationary Pallas GPU kernel, the TPU kernel tiles receiver-ordered edges into blocks of $B$. For each tile, VMEM holds the edge-local operands and messages, while the TensorCore's SMEM holds the corresponding sender and receiver indices. The kernel copies ${\bf y}_t$ and ${\bf s}_t$ from HBM to VMEM, loads ${\bf a}_t$ and ${\bf b}_t$ into SMEM, gathers the sender features $\hat{\bf x}_t={\bf x}[{\bf a}_t]$ into VMEM, and computes the edge-level weighted messages ${\bf \tilde m}_{ab}$ of Eq.~\ref{eq:message-passing}. These messages are then reduced over receivers to form ${\bf m}_b=\sum_{a\sim b}{\bf \tilde m}_{ab}$.

The reduction state is retained across consecutive tiles, so a receiver spanning a tile boundary is accumulated before being written to HBM (Algorithm~\ref{alg:tpu-conv}). Like the CUDA and Pallas GPU convolution kernels, this fuses message formation with aggregation and therefore avoids materializing the complete edge-leading message tensor in HBM. The TPU-specific distinction is the edge-tiled schedule: VMEM holds a block of edge messages, while SMEM carries the edge indices used for the gathers and segmented receiver reduction.

\begin{algorithm}[ht]
\caption{Fused message-passing convolution on TPU.}
\label{alg:tpu-conv}
\DontPrintSemicolon
$\mathit{acc} \gets 0$, $\mathit{current} \gets -1$ \;
\For{\textnormal{each tile} $t$ \textnormal{of} $B$ \textnormal{receiver-ordered edges}}{
  ${\bf y}_t,{\bf s}_t \gets$ HBM to VMEM,
  \quad ${\bf a}_t,{\bf b}_t \gets$ HBM to SMEM \;
  $\hat{\bf x}_t \gets$ HBM to VMEM
  ${\bf x}[{\bf a}_t[k]]$, $k=0,\ldots,B-1$ \;
  ${\bf \tilde m}_t
  \gets \textsc{MessageTile}(\hat{\bf x}_t,{\bf y}_t,{\bf s}_t)$ \;
  $\textsc{ReduceFlush}({\bf b}_t,{\bf \tilde m}_t,
  \mathit{acc},\mathit{current})$ \;
}
$\textsc{Flush}(\mathit{current})$
\end{algorithm}

\paragraph{Coefficient packing and Clebsch--Gordan contraction.}
\textsc{MessageTile} evaluates the contraction path by path. The static records $(i0_p,i1_p,i2_p,c_p)$ are grouped on the host first by output coordinate $i0_p$ and then by l.h.s.\ input coordinate $i1_p$. This allows each $B\times N_q$ tile $\hat{\bf x}^{i1_p}$ to be loaded once and reused across the corresponding $i2_p$ paths (Algorithm~\ref{alg:tpu-msg}).

As in the Pallas GPU kernel, the sparse coefficient loops are unrolled at trace time and $c_p$ is embedded directly in the generated computation. Unlike Algorithm~\ref{alg:tp-trailing} for CUDA, the TPU kernel therefore does not load and interpret coefficient records at run time.

\paragraph{Factoring out the edge scalar.}
The edge scalar $s^{i3(i0),q}_{ab}$ in Eq.~\ref{eq:message_passing_convolution} is the same for every path of output coordinate $i0$, so the kernel pulls it out of the path sum, as in Eq.~\ref{eq:message-passing}:
\begin{equation*}
  \tilde m^{i0,q}_{ab}
  = s^{i3(i0),q}_{ab}\,\underbrace{\sum_{i1,i2} C^{i0}_{i1,i2}\,x^{i1,q}_a\,y^{i2,q}_{ab}}_{z^{i0,q}_{ab}}.
\end{equation*}
It accumulates $z$ first, then multiplies by the edge scalar once per output coordinate instead of once per path.

\begin{algorithm}[ht]
\caption{\textsc{MessageTile}}
\label{alg:tpu-msg}
\DontPrintSemicolon
\For{$i0$ \textnormal{ output coordinate}}{
  $z \gets 0$ \;
  \For{$i1$ \textnormal{ l.h.s.\ input coordinate of } $i0$}{
    \For{$(i2_p,c_p)$ \textnormal{ path of } $(i0,i1)$}{
      $z
      \mathrel{+}=
      c_p\,\hat{\bf x}[i1]\,{\bf y}[i2_p]$
      \tcp*{contraction before mixing}
    }
  }
  ${\bf \tilde m}[i0]
  \gets
  z\,{\bf s}[i3(i0)]$
  \tcp*{scalar mixing}
}
\end{algorithm}

\paragraph{Reducing over receivers.}
\textsc{ReduceFlush} performs a segmented sum of the edge-message tile ${\bf \tilde m}_t$, of shape $(D,B,N_q)$. Because the edges are ordered by receiver, equal receiver indices form contiguous segments. The routine walks the $B$ edges eight at a time, accumulates messages belonging to $\mathit{current}$, and flushes the completed receiver sum to ${\bf m}_{\mathit{current}}$ when the receiver changes (Algorithm~\ref{alg:tpu-reduce}). The variables $\mathit{acc}$ and $\mathit{current}$ persist across tiles.

\begin{algorithm}[ht]
\caption{\textsc{ReduceFlush}}
\label{alg:tpu-reduce}
\DontPrintSemicolon
\SetKwProg{Fn}{Function}{:}{}
\SetKwFunction{ReduceFlush}{ReduceFlush}
\SetKwFunction{Flush}{Flush}
\Fn{\ReduceFlush{${\bf b}_t$, ${\bf \tilde m}$, $\mathit{acc}$, $\mathit{current}$}}{
  \For{\textnormal{each chunk of} $8$ \textnormal{edges, receivers} $b[0,\ldots,7]$}{
    $\mathit{start} \gets 0$ \;
    \tcp{entered when the chunk crosses a receiver boundary}
    \If{$b[0] \neq \mathit{current}$ \textnormal{\textbf{or}} $b[0] \neq b[7]$}{
      \For{$j=0,\ldots,7$ \textnormal{ with } $b[j]\neq b[j-1]$,
      \ $b[-1]\equiv\mathit{current}$}{
        $\mathit{acc}
        \mathrel{+}=
        {\bf \tilde m}[\,\mathit{start}\leq\text{edge}<j\,]$
        \tcp*{complete current receiver}
        \lIf{$\mathit{current}\geq0$}{\Flush{$\mathit{current}$}}
        $\mathit{current}\gets b[j]$,
        \quad $\mathit{start}\gets j$ \;
      }
    }
    $\mathit{acc}
    \mathrel{+}=
    {\bf \tilde m}[\,\text{edge}\geq\mathit{start}\,]$ \;
  }
}
\BlankLine
\Fn{\Flush{$b$}}{
  $\mathit{stage}
  \gets\sum_{\text{sublanes}}\mathit{acc}$,
  \quad
  $\mathit{acc}\gets0$ \;
  VMEM to HBM
  $\mathit{stage}\to{\bf m}_b$ \;
}
\end{algorithm}

\paragraph{Backward.}
The backward follows the same edge-tiled organization as Algorithm~\ref{alg:tpu-conv}. In addition to the edge-local ${\bf y}_t$ and ${\bf s}_t$, it gathers ${\bf x}_{a}$ and the receiver cotangent $d{\bf m}_b$ for each edge, replaces \textsc{MessageTile} by the sweep in Algorithm~\ref{alg:tpu-bwd}, and reduces the contributions to $d{\bf x}_a$ using \textsc{ReduceFlush} keyed on senders. This segmented reduction therefore uses a sender-ordered edge traversal. The cotangents $d{\bf y}_{ab}$ and $d{\bf s}_{ab}$ remain edge-local.

The sweep makes one pass over the same static coefficient records to compute the three cotangents of Eq.~\ref{eq:conv-bwd}. Since $\tilde m^{i0,q}_{ab}=s^{k,q}_{ab}\,z^{i0,q}_{ab}$ with $k=i3(i0)$, the edge scalar and the contraction also separate in the backward:
\begin{equation*}
  w^{i0,q}_{ab} = s^{k,q}_{ab}\,dm^{i0,q}_b,
  \qquad
  ds^{k,q}_{ab} = \sum_{i0\,:\,i3(i0)=k} z^{i0,q}_{ab}\,dm^{i0,q}_b.
\end{equation*}
The sweep computes $w$ once per output coordinate and uses it in both the $d{\bf x}$ and $d{\bf y}$ updates. The edge-scalar cotangent instead needs $z$, which the sweep recomputes from the same paths. Because the records are grouped by block $k$, each $d{\bf s}[k]$ is accumulated in registers and written once. The r.h.s.\ ${\bf y}_{ab}$ is shared across channels, so its cotangent is summed over $q$ at the end.

\begin{algorithm}[ht]
\caption{Backward sweep of one edge tile.}
\label{alg:tpu-bwd}
\DontPrintSemicolon
$d\hat{\bf x},d{\bf y}\gets0$ \;
\For{$k$ \textnormal{ scalar-mixing block}}{
  $\sigma\gets0$ \;
  \For{$i0$ \textnormal{ output coordinate of block } $k$}{
    $w
    \gets
    {\bf s}[k]\,d{\bf m}[i0]$
    \tcp*{mix the cotangent once}
    $z\gets0$ \;
    \For{$(i1_p,i2_p,c_p)$ \textnormal{ path of } $i0$}{
      $z
      \mathrel{+}=
      c_p\,{\bf y}[i2_p]\,\hat{\bf x}[i1_p]$
      \tcp*{as in the forward}
      $d\hat{\bf x}[i1_p]
      \mathrel{+}=
      c_p\,{\bf y}[i2_p]\,w$ \;
      $d{\bf y}[i2_p]
      \mathrel{+}=
      c_p\,\hat{\bf x}[i1_p]\,w$ \;
    }
    $\sigma
    \mathrel{+}=
    z\,d{\bf m}[i0]$ \;
  }
  $d{\bf s}[k]\gets\sigma$
  \tcp*{one write per block}
}
$d{\bf y}\gets\sum_q d{\bf y}$ \;
\end{algorithm}

\paragraph{VMEM budget.}
Each buffer has at most three axes: the tile's $B$ edges, the $N_q$ channels, and the feature components of a node, edge, or message value, whose count grows with $\lmax$. The forward holds the gathered node features and message tile, one edge-scalar array per irreducible-representation block, and the segmented sum accumulator together with the row used for flushing; sender and receiver indices are held in the TPU TensorCore's SMEM. The backward additionally holds the three cotangents and two copies of ${\bf x}$ and $d{\bf m}$ for pipelining.

Together these buffers occupy $4.3$~MiB in the forward and $5.4$~MiB in the backward at $\lmax=3$, $N_q=128$, and fp32, and roughly half as much at $\lmax=2$ for the same block sizes.

In addition to these buffers, the unrolled contraction of Algorithm~\ref{alg:tpu-msg} uses VMEM for compiler spills from vector registers, taking the peak to $13.1$ and $10.1$~MiB of the TPU TensorCore's $16$~MiB budget (Table~\ref{tab:tpu-vmem}).

\begin{table}[ht]
\centering
\caption{VMEM usage on a TPU v4 TensorCore of the fused TPU kernel at $\lmax = 3$, $C = 128$, fp32, in MiB against the $16$~MiB per-core budget; \emph{buffers} counts the kernel's own arrays, \emph{peak VMEM} adds the compiler's spill.}
\label{tab:tpu-vmem}
\vspace{0.4em}
\begin{tabular}{lccc}
\toprule
pass & $B$ & buffers & peak VMEM \\
\midrule
fwd & 128 & $4.3$ & $13.1$ \ ($82\%$) \\
bwd &  32 & $5.4$ & $10.1$ \ ($63\%$) \\
\bottomrule
\end{tabular}
\end{table}

\section{Relevant GPU and TPU architectural concepts} \label{apx:hardware_concepts}

In order to provide a more comprehensive description of our kernels, we begin with an overview of the relevant hardware concepts for both GPU and TPU, as well as their differences. This section is not intended as an exhaustive or even fully accurate description of these respective devices, but instead as a structured glossary of concepts useful to understand later sections.  

\subsection{GPU and TPU accelerators comparison}

GPUs are massively parallel architectures exposing a single-instruction, multiple-threads (SIMT) programming model that gives fine-grained control over every level of parallelism via the CUDA \Cxx\, language extension \citep{CUDAC++}. A typical server-grade GPU embeds about a hundred streaming multi-processors (SMs) on the same die, each SM being capable of scheduling 2048 threads and carrying a low-latency 256 KB memory bank (L1 cache) alongside a 256 KB register file.

In contrast, TPUs rely on the Mosaic compiler to lower abstract array operations onto dedicated processing units, including a systolic matrix multiply unit (MXU) and single-instruction, multiple-data (SIMD) vector processing unit (VPU) \citep{TPU-Jouppi17}. A TPU tray typically consists of 4 TPU chips, each containing one or two cores, each backed by large (tens of MB) dedicated memory banks. 

All these architectural differences translate into significant algorithmic differences between the optimized kernels on each platform, on which more details can be found in the sections below. In particular, since end-to-end performance can depend strongly on reducing memory traffic and avoiding the materialization of intermediate results through operation fusion, different hardware architectures may opt for different fusion strategies and very different sizes for temporary buffers.

The algorithmic development of {\tt e3j}, as outlined in appendix \ref{apx:kernel_description} reflects these differentiations.

Furthermore, significant engineering effort has been recently made to optimize 
or diversify the compiler stack. Just-in-time (JIT) compilation, from either CUDA or domain-specific languages embedded in Python such as Pallas, allows the production of specialized device assembly code from static problem parameters. It is now an ubiquitous alternative to traditional ahead-of-time (AOT) compilation of CUDA source into SASS or PTX. Further details on these distinctions can be found in appendix \ref{sec:pallas-cuda}. 

\subsection{Overview of relevant GPU concepts}

\paragraph{Computational units.}
A GPU exposes a hierarchical execution model that maps a large number of software threads onto parallel hardware resources. It is useful to distinguish the programming hierarchy: threads, warps and cooperative thread arrays (CTAs), from the underlying hardware hierarchy of streaming multiprocessors (SMs) and their execution units.
\begin{itemize}
    \item At the hardware level, the GPU is composed of many \textbf{streaming multiprocessors} (SMs). Each SM contains several execution pipelines for floating-point and integer arithmetic, load/store operations, specialized functions, and matrix operations such as those implemented by Tensor Cores.
    \item At the software level, a \textbf{thread} is one logical execution instance of the kernel program. Each thread maintains its own execution state, including registers and, on modern NVIDIA GPUs, its own program counter.
    \item Threads are organized in CTA, which are partitioned into groups of 32 threads called \textbf{warps}. An SM schedules and issues instructions at warp granularity: typically, one instruction is issued to the active threads of a warp, which apply that instruction to their respective data. This execution model is referred to as \textbf{Single Instruction, Multiple Threads} (SIMT). Threads may follow different control-flow paths, although divergence within a warp generally reduces execution efficiency.
    \item A \textbf{cooperative thread array} (CTA), also called a CUDA thread block, groups threads that cooperate on the same unit of work. All threads of a CTA are scheduled on the same SM, where they may synchronise and exchange data through shared memory. The SM partitions the CTA into warps and interleaves the execution of its ready warps. Multiple CTAs may reside concurrently on one SM when register and shared
    memory capacity permit.
\end{itemize}

\paragraph{Memory hierarchy.}
GPU performance is strongly influenced by where data reside in the memory hierarchy. Storage closer to the execution units provides high bandwidth and low latency but limited capacity, motivating kernels to move data through successively smaller on-chip memories and maximize reuse before returning to global memory.
\begin{itemize}
    \item \textbf{Registers} provide the storage closest to arithmetic execution. Registers hold thread-local values such as operands, intermediate results and accumulators, and are allocated from an SM-wide    register file among the resident threads. On H100, each SM provides $65\,536$ 32-bit registers, corresponding to $256$~KiB of register-file capacity. Their limited capacity makes register usage an important
    constraint: using more registers per thread or CTA can reduce the number of CTAs that can reside concurrently on an SM.

    \item \textbf{Shared memory (SMEM)} is an explicitly managed on-chip scratchpad shared by the threads of a CTA. It enables fast data reuse and communication between cooperating threads. The H100 provides a combined $256$~KiB L1/texture-cache and shared-memory pool per SM, of which up to $228$~KiB can be configured as shared memory. Although L1 cache and SMEM use the same underlying capacity, they have different programming semantics: L1 is hardware-managed cache, whereas placement and access in SMEM are controlled explicitly by the kernel.

    \item The \textbf{L2 cache} is a larger cache shared across the GPU and forms the last on-chip caching level before device memory. H100 GPUs provide a $50$~MB L2 cache, allowing frequently accessed data to be
    retained on-chip and reducing repeated accesses to the substantially larger off-chip memory.

    \item \textbf{Global memory (GMEM)} is the GPU-wide memory address space used to store the large tensors processed by a kernel. Its backing storage is off-chip high-bandwidth DRAM (HBM3 on the H100 SXM5, for example) and consequently has much greater capacity but higher access cost than registers, SMEM or the on-chip caches.  Efficient kernels therefore aim to minimize global-memory traffic and to reuse data after
    bringing it on-chip.

    \item On Hopper GPUs, the \textbf{Tensor Memory Accelerator (TMA)} provides a specialized mechanism for asynchronously transferring multidimensional blocks of data between GMEM and SMEM. Once a transfer
    has been initiated, computation may proceed independently while TMA moves the next block of data. Software pipelines can consequently overlap global-memory traffic with arithmetic, as exploited by the Mosaic GPU
    kernel described below.
\end{itemize}

\paragraph{Performance limits.}
For many ML kernels, particularly those with low arithmetic intensity, the rate at which data can be transferred between HBM and the GPU is the dominant performance bottleneck. In this memory-bound regime, the peak HBM bandwidth provides a useful ``speed-of-light'' estimate: the minimum execution time is bounded by the number of bytes that must be transferred to and from HBM divided by the maximum sustainable HBM bandwidth.  Kernel efficiency can therefore be assessed by comparing its achieved HBM bandwidth with the hardware peak. Kernels with sufficiently high arithmetic intensity may instead become compute-bound, in which case peak arithmetic throughput rather than HBM bandwidth determines the relevant performance ceiling. 

\subsection{Overview of relevant TPU concepts}

\paragraph{Computational units.}
A TPU exposes a different execution model from the SIMT hierarchy of a GPU. Rather than scheduling many independent software threads, a TPU TensorCore is programmed approximately as a sequential machine operating on wide vector tiles.
\begin{itemize}
    \item A TPU v6e chip contains one \textbf{TensorCore}, comprising a \textbf{scalar unit}, a \textbf{vector processing unit} (VPU), and two \textbf{matrix-multiply units} (MXUs). The scalar unit handles control,
    indexing and scalar arithmetic, the VPU performs general vector operations, and the MXUs provide high-throughput matrix multiplication.

    \item The natural unit of vector computation is a two-dimensional \textbf{vector-register tile}. For 32-bit values on TPU v6e, registers are organized as $8\times128$ elements (sublanes $\times$ lanes).     Vector instructions operate on these tiles collectively rather than on independently scheduled threads.

    \item Array layout is therefore closely tied to this native tile shape. Operations are most efficient when trailing dimensions map cleanly onto the $8\times128$ register layout; poorly aligned or very small arrays may waste vector capacity or require additional rearrangement.

    \item The two \textbf{MXUs} are specialized systolic units for dense matrix multiplication, while the VPU handles the more general elementwise,
    reduction and permutation operations used throughout Pallas kernels.
\end{itemize}

\paragraph{Memory hierarchy.}
Large tensors reside in off-chip HBM and are staged through software-managed on-chip memories before computation.
\begin{itemize}
    \item \textbf{Vector registers (VREGs)} hold operands and intermediate array-valued results consumed by the VPU and MXUs. Their limited capacity makes register pressure and spilling important considerations.

    \item \textbf{Vector memory (VMEM)} is a large software-managed on-chip scratchpad for array data. TPU v6e provides approximately $128$~MiB of VMEM per TensorCore, allowing substantially larger working sets to remain on-chip than in GPU shared memory.

    \item \textbf{Scalar memory (SMEM)} is a separate on-chip memory for scalar data such as indices and control information. TPU v6e provides approximately $1$~MiB of SMEM. This should not be confused with GPU ``SMEM'', which denotes shared memory.

    \item \textbf{HBM} provides the large off-chip storage for kernel inputs and outputs. TPU v6e provides $32$~GB of HBM with approximately $1.64$~TB/s peak bandwidth. Data are transferred between HBM and VMEM
    using asynchronous DMA engines, allowing future tiles to be prefetched while the current tile is being processed.
\end{itemize}

\paragraph{Performance limits.}
As on GPUs, low-arithmetic-intensity TPU kernels may be bounded by HBM bandwidth, whereas compute-intensive kernels may instead be limited by VPU or MXU throughput. Efficient TPU kernels therefore aim to retain data in VMEM, respect the native $8\times128$ vector layout, and overlap HBM--VMEM transfers with computation.

\subsubsection{Relevant distinction between GPU and TPU architectures}

The GPU and TPU kernels are shaped by fundamentally different execution and memory models. On a GPU, computation is organized around CTAs composed of SIMT warps, with thread-local registers and a relatively small CTA-local shared-memory scratchpad. By contrast, a TPU TensorCore operates on wide $8\times128$ vector-register tiles and provides a much larger software-managed VMEM working memory. On H100, at most $228$~KiB of shared memory is available per SM, whereas TPU v6e provides approximately $128$~MiB of VMEM per TensorCore. These differences lead naturally to different kernel organizations: the GPU kernel keeps a receiver reduction local to one CTA and streams edge data through SMEM, whereas the TPU kernel operates on larger VMEM-resident windows using tile-wide vector operations.

\section{Reconstructions} \label{apx:reconstructions}

\subsection{Reconstruction of harmonic polynomials.} 
\label{apx:harmonics}

The basis $(Y_l^m)_{-l\leq m \leq l}\in\Y_l$ of spherical harmonic polynomials of order $l$ depends on the basis $(x, y, z)$ of $\R^3$ by the spin equations
demanding that $Y_l^m$ is an eigenvector of $z$-axis (infinitesimal) rotations,
\begin{equation} \label{eq:sigma-z-ylm}
    \sigma_z \cdot Y_l^m = i m Y_l^m
\end{equation}
Noting that $\sigma_z$ can be written as the infinitesimal rotation 
$r \partial_\phi$ of the longitude angle $\phi$, equation \eqref{eq:sigma-z-ylm}
implies that $Y_l^m$ varies as ${\rm e}^{im\phi}$ along
the longitude angle $\phi$. Up to a choice of normalization factor $c_l$,
for $m = \pm l$, this implies that $Y_l^{\pm l}$ is the fastest 
oscillating degree $l$-monomial with respect to $\phi$:
\begin{equation}
    Y_l^{\pm l}(x, y, z) = c_l \, (x + iy)^l= 
    c_l \, r^l \cos(\theta)^l  \, {\rm e}^{\pm il\phi}.
\end{equation}
In general, polynomials of lower absolute spin 
$Y_l^m$ with $|m|< l$ similarly take the form 
\begin{equation}
    Y_l^m (x, y, z) =  r^l P_{lm}(\cos \theta, \sin \theta) \, {\rm e}^{im\phi},
\end{equation}
and is an eigenvector of $\sigma_z$ with eigenvalue $i m$ 
(in particular, $Y_l^0$ is always invariant with respect to $z$-axis rotations).
They can be obtained from $Y_l^{\pm l}$ by iterating
the ladder operators $\sigma_\pm = \sigma_x \pm i \sigma_y$, which increase / decrease the magnetic 
quantum number $m$ by 1.

Note that our choice of generators 
is in agreement with physicists and chemists' quantum state 
notation $|lm\rangle$, while real harmonics $Y_{lm}$ are not eigenvalues 
of $\sigma_z$ : only $|m|$ is determined.

In practice, the computation of polynomials $Y_l^m$ is cached and 
staged out of XLA compilation. 
It returns an integer-valued array of exponents ${\tt Y.exp} :: [M, 3]$, 
and a sparse coefficient matrix ${\tt Y.coef} :: [N, M]$ 
where $N$ is the number of polynomials in the current basis, 
and $M$ the number of distinct monomials required for their evaluation. 

\subsection{Reconstruction of Clebsch Gordan coefficients}

The irreducible CG array
$C^{LM}_{lm,l'm'}$ can be reconstructed from 
$C^{LL}_{ll,l'l'}$, 
expressing the top-spin eigenvectors with respect 
to the pure generators.
The total $z$-spin $M = m + m'$ can indeed be lowered by 
the ladder operator $\sigma_-$, acting as 
$\sigma_- \otimes 1 + 1 \otimes \sigma_-$ on the 
tensor product space by the Leibniz rule, 
and any eigenvector $|LM\rangle$ for $M < L$ can be obtained up to 
a normalization factor by iterating $\sigma_-$ from $|LL\rangle$.

Therefore one only need to solve the eigenvalue equations 
$\vec\sigma^2 |LL\rangle = L(L+1) |LL\rangle$ 
and $\sigma_z |LL\rangle = L |LL\rangle$ to obtain the top-spin 
generators of each irreducible subspace of $\Y_l \otimes \Y_{l'}$, 
for each value of $L \in \{ |l-l'|,\dots, l+l'\}$.
The eigenvalue equations can be expressed and easily 
solved as a triangular system, with the algorithm documented below. 

By the second order Leibniz rule, the operator $J_+ J_-$ acts on a dyadic
tensor product as:

\begin{equation}
        J₊J₋  =  (J₊J₋⊗ 1) + (1⊗ J₊J₋)
                + (J₊ ⊗ J₋) + (J₋⊗ J₊)
\end{equation}

Since $J^2 = J_+ J_- + J_z + J_z^2$, 
the action of squared angular momentum
on a pure state $|lm,l'm'⟩$ is given by

\begin{equation}
\begin{split}
J²|lm,l'm'⟩ &= \big\{ l(l+1) + l'(l'+1) + 2mm' \big\} \cdot |lm,l'm'⟩ \\
            &+ c₋(m)c'₊(m') \cdot |l(m-1),l'(m'+1)⟩ \\
            &+ c₊(m)c'₋(m') \cdot |l(m+1),l'(m'-1)⟩,
\end{split}        
\end{equation}
where the term in curly brackets contains the independent action of 
$J^2$ on each operand, and twice the action of $J_z \otimes J_z$. 

Assuming $V = \sum V_{m,m'} |lm,l'm'⟩$ is an eigenvector $|LL⟩$ for some $L$, 
by additivity of the $z$-spin $M=L$ we can write more succinctly:
\begin{equation}
V = \sum_m V_m \cdot |lm,l'(L-m)⟩ = \sum_m V_m \cdot |m,(L-m)⟩
\end{equation}
so that the eigenvalue equation reads as follows:
\begin{equation}
\begin{split}
L(L+1) \cdot V_m &= \{l(l+1) + l'(l'+1) + 2mm'\} \cdot V_m \\
    &+ c-(m+1)c'+(m'-1) \cdot V_{m+1} \\
    &+ c+(m-1)c'-(m'+1) \cdot V_{m-1}.
\end{split}
\end{equation}
Given $m$ is constrained to $[-L, L]$, coordinates $V_m$ of 
the max-spin eigenvector $|LL\rangle$ can easily be constructed 
by solving the above triangular system in dense or sparse format.

%\input{content/A_groups}
%\input{content/B_ladder_operators}
%\input{content/C_clebsch_gordan}
%\input{content/D_schur_lemma}

%%%%%%%%%%%%%%%%%%%%%%%%%%%%%%%%%%%%%%%%%%%%%%%%%%%%%%%%%%%%

\clearpage
\newpage

\end{document}